\documentclass[runningheads]{llncs}

\usepackage{eccv}

\usepackage{eccvabbrv}

\usepackage{graphicx}
\usepackage{booktabs}

\usepackage[accsupp]{axessibility}  

\usepackage{hyperref}

\usepackage{orcidlink}

\usepackage{tabularx}
\usepackage{multirow}
\usepackage{multicol}
\usepackage{graphicx}
\usepackage{caption}
\usepackage{amsmath}
\usepackage{amssymb}

\usepackage{xspace}
\makeatletter
\DeclareRobustCommand\onedot{\futurelet\@let@token\@onedot}
\def\@onedot{\ifx\@let@token.\else.\null\fi\xspace}
\def\eg{\emph{e.g}\onedot} 
\def\ie{\emph{i.e}\onedot} 
 
\def\etc{\emph{etc}\onedot} 
 
\def\etal{\emph{et al}\onedot}
\makeatother

\usepackage{array}
\newcolumntype{H}{>{\setbox0=\hbox\bgroup}c<{\egroup}@{}}
\usepackage[export]{adjustbox}
\usepackage{multirow}
\usepackage{todonotes} 

\usepackage{cuted}
\usepackage{capt-of}

\usepackage[normalem]{ulem}
\useunder{\uline}{\ul}{}

\usepackage{booktabs}  

\usepackage[font=scriptsize]{caption}
\usepackage[font=scriptsize]{subcaption}

\begin{document}

\title{Weakly-supervised Camera Localization by Ground-to-satellite Image Registration} 

\titlerunning{Weakly-supervised Ground-to-satellite Camera Localization}


\author{Yujiao Shi\inst{1}\orcidlink{0000-0001-6028-9051} \and
Hongdong Li\inst{2}\orcidlink{0000-0003-4125-1554} \and
Akhil Perincherry \inst{3}\orcidlink{0009-0001-1713-2494} \and
Ankit Vora \inst{3} \orcidlink{0000-0001-7976-8730}}

\authorrunning{Shi et al.}

\institute{ShanghaiTech University, China \and
The Australian National University, Australia \and
Ford Motor Company, USA\\
\email{shiyj2@shanghaitech.edu.cn}}

\maketitle

\begin{abstract}
The ground-to-satellite image matching/retrieval was initially proposed for city-scale ground camera localization. 
This work addresses the problem of improving camera pose accuracy by ground-to-satellite image matching after a coarse location and orientation have been obtained, either from the city-scale retrieval or from consumer-level GPS and compass sensors. 
Existing learning-based methods for solving this task require accurate GPS labels of ground images for network training. 
However, obtaining such accurate GPS labels is difficult, often requiring an expensive {\color{black}Real Time Kinematics (RTK)} setup and suffering from signal occlusion, multi-path signal disruptions, \etc. 
To alleviate this issue, this paper proposes a weakly supervised learning strategy for ground-to-satellite image registration when only noisy pose labels for ground images are available for network training.
It derives positive and negative satellite images for each ground image and leverages contrastive learning to learn feature representations for ground and satellite images useful for translation estimation. 
We also propose a self-supervision strategy for cross-view image relative rotation estimation, which trains the network by creating pseudo query and reference image pairs. 
Experimental results show that our weakly supervised learning strategy achieves the best performance on cross-area evaluation compared to recent state-of-the-art methods that are reliant on accurate pose labels for supervision.
 
       
\end{abstract}


\section{Introduction}

Camera localization is pivotal in real-world applications such as autonomous driving, field robotics, and augmented/virtual reality. Recent research has explored various methods to approximate the coarse location and orientation of a ground camera, such as visual retrieval techniques, consumer-level GPS and compass, \etc.  
To attain greater pose accuracy, other sensors like Lidar~\cite{vora2020aerial,mishra2022infra,veronese2015aerial}, Radar~\cite{tang2020rsl,tang2021self}, and High Definition (HD) maps, have been investigated. However, many commercial autonomous vehicles at level two/three lack these sensors. Maintaining and updating high-precision HD maps~\cite{xiao2020monocular,liu2020high} is challenging and expensive. In response, satellite imagery has emerged as an inexpensive alternative reference source due to its wide accessibility and global coverage.

We focus on ground-to-satellite camera localization, aiming to determine a ground camera's location and orientation relative to a geo-tagged satellite map. Prior research~\cite{workman2015location,vo2016localizing,Hu_2018_CVPR,Liu_2019_CVPR,zhai2017predicting,Regmi_2019_ICCV,shi2019spatial,toker2021coming,Zhu_2022_CVPR,DeBNet} has centered on city-scale camera localization, employing image retrieval to match ground and satellite images. However, image retrieval introduces errors that can span tens of meters. Recent efforts have addressed this by enhancing camera pose accuracy through ground-to-satellite image registration, guided by coarse location and orientation estimates.  Nonetheless, the significant viewpoint differences between ground and satellite images make handcrafted features fail~\cite{shi2022accurate}. Learning-based approaches~\cite{zhu2021vigor,xia2022visual,shi2022beyond,lentsch2023slicematch,fervers2022uncertainty} require a large training dataset with accurate ground truth (GT) poses for the ground images. 
This often requires field surveys or the use of specialized equipment, which can be costly and time-consuming, limiting the availability of GT data, particularly for large-scale or extensive image datasets.
Furthermore, even the expensive high-accuracy RTK GPS signals can be affected by factors such as multipath interference, signal blockage, and atmospheric conditions, leading to inaccuracies in location data~\cite{Geiger2013IJRR,maddern20171}.
Therefore, this paper aims to develop a weakly supervised ground-to-satellite image registration strategy to increase the ground cameras' pose accuracy when only coarse pose labels for ground images are provided.

\begin{figure}[t]
    \centering
        \setlength{\abovecaptionskip}{0pt}
    \setlength{\belowcaptionskip}{0pt}
    \adjincludegraphics[width=\linewidth,trim={{0\width} {0.0\width} {0\width} {0.0\width}},clip]{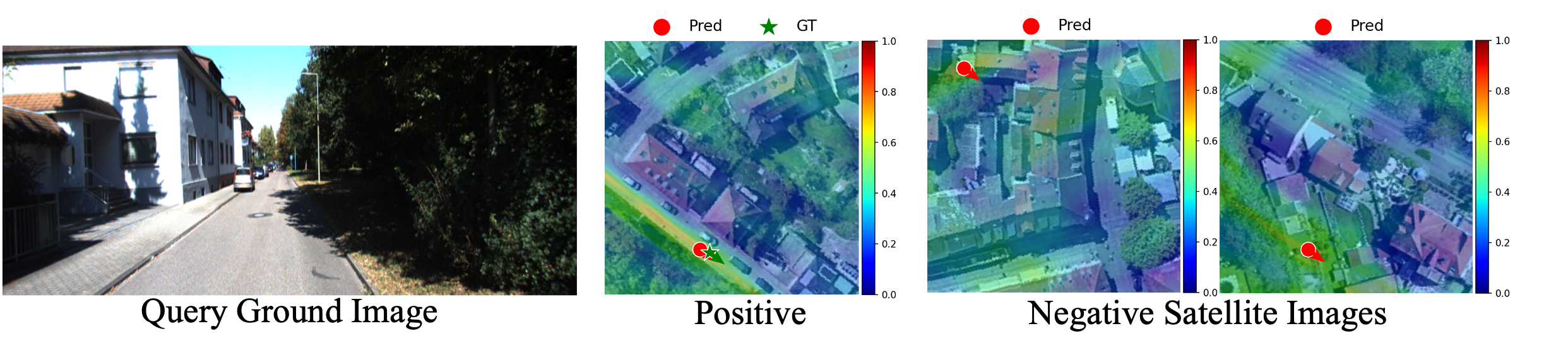}
    \caption{\scriptsize We derive positive and negative satellite images for each ground-view image based on its coarse location information. A similarity map between the ground view and each satellite image is computed when aligned at different locations. Our training objective aims to maximize the maximum similarity in the positive similarity map while minimizing the maximum similarity in the negative similarity map.}
    \label{fig:openfigure}
\end{figure}

We target the 3-DoF (degree-of-freedom) pose estimation for ground cameras, \ie, 2-DoF location and 1-DoF orientation. 
Under the weakly-supervised setting, deterministic pose output by a network is suboptimal, as no GT pose is available to supervise the network output. 
We resort to the recent representation learning approaches by contrastive learning to solve this problem.

Given the coarse pose of a ground camera, we determine a satellite image that covers the local surroundings of the camera and some satellite images that do not cover the local surroundings, which are regarded as positive and negative for this ground image, respectively. 
We utilize the signal that a ground image is within its positive satellite image while outside its negative satellite images to train a network.
The network is trained to learn feature representations such that the similarity between the ground and its positive satellite images at their optimal relative pose is more significant than between the ground and its negative satellite images at their ``optimal'' relative pose.

The ``optimal'' relative pose for a ground-and-satellite image pair is determined by the maximum cross-view similarity when aligned at this relative pose (among others). We follow the standard synthesis-and-matching procedure to determine this ``optimal'' relative pose: first synthesizing an overhead-view feature map from the ground view image and then matching it against the reference satellite feature map. 
Conventional methods~\cite{fervers2022uncertainty,sarlin2023orienternet} for this matching process rotate and translate the synthesized overhead view feature map according to predetermined candidate poses. Considering the vast search space for the 3-DoF pose, we decouple the rotation and translation estimation and develop different supervision strategies for them, respectively, following Shi et al.~\cite{shi2023boosting}.

The rotation estimation is conducted first, and the framework is designed by network regression. It takes input as the query and reference images and outputs their relative pose. To supervise this network, we create some ``satellite and satellite'' image pairs with GT relative poses. 
Specifically, we randomly rotate and translate a satellite image. The transformed satellite image mimics a synthesized overhead view image from a ground image. 
We train the network to estimate the relative pose between the original and transformed satellite images. 
Once the network is trained, it is leveraged to estimate the relative orientation between a query ground image and its positive satellite image.

After this, our translation estimation framework synthesizes an overhead-view feature map from the ground image with the orientation aligned with satellite images and then matches it against a satellite feature map. The output is a similarity (/location probability) map of the ground image with respect to the satellite image.  
This process is implemented between a query image and its positive and negative satellite images,  
as illustrated in Fig.~\ref{fig:openfigure}. A contrastive learning supervision strategy is employed to train the network by maximizing the maximum similarity in the positive similarity map while minimizing the maximum similarity in the negative similarity map.

We conduct experiments on  KITTI~\cite{Geiger2013IJRR}, a popular autonomous driving dataset where ground images are captured by a pin-hole camera with limited field-of-view (FoV), and VIGOR~\cite{zhu2021vigor}, a well-known cross-view localization dataset where ground images are panoramas. 
Experimental results demonstrate that our method achieves the best generalization ability compared to the recent state-of-the-art despite not requiring accurate pose labels for supervision.

Our contributions are summarized as follows:
\begin{itemize}
    \item We propose a self-supervised relative rotation estimation strategy between ground-and-satellite image pairs;
    \item We introduce a weakly-supervised translation estimation strategy for ground-to-satellite camera pose estimation;
    \item Our method achieves the best cross-area generalization ability than the previous fully supervised state-of-the-art that relies on GT labels for training. 
\end{itemize}

\section{Related Work}

\subsection{Cross-view image-based localization}

\textbf{City-scale localization. }
Ground-to-satellite image-based localization aims to determine the location of a ground camera by matching it with a satellite map covering the region of interest. Initially proposed for city-scale coarse-level localization, it was formulated as an image retrieval problem. Specifically, a large satellite map of the interested region is first split into small patches to construct a geo-referenced satellite image database. For a query image captured at the ground level, its similarity with every database satellite image is computed, and the GPS of the most similar satellite image is taken as the query camera's location.
Over the past decades, hand-crafted features~\cite{castaldo2015semantic,lin2013cross,mousavian2016semantic} have been demonstrated to be a bottleneck for cross-view feature matching due to significant geometric and appearance variations. Seminal works using deep networks~\cite{workman2015location,workman2015wide,vo2016localizing} demonstrated that the learned feature descriptors by a metric learning training objective offer better and more reliable performance. Researchers have since investigated learning powerful feature descriptors~\cite{Cai_2019_ICCV,yang2021cross,Zhu_2022_CVPR,GeoText1652}, orientation-aware cross-view representations~\cite{Hu_2018_CVPR,Liu_2019_CVPR,sun2019geocapsnet,zhu2021revisiting}, and different strategies for bridging the cross-view domain gap~\cite{zhai2017predicting,Regmi_2019_ICCV,shi2019spatial,shi2020optimal,toker2021coming}. Cross-view localization has also been extended from a 2-DoF location estimation task to a 3-DoF joint location and orientation estimation task~\cite{shi2020looking}, and from a single image-based localization problem to a video-based localization problem~\cite{vyas2022gama,shi2023cvlnet,zhang2023cross}.
However, the database images are often discretely sampled, while query images' locations are continuous. Thus, the image retrieval formulation results in a coarse camera pose estimation. Its accuracy is determined by the sample density of ground images.

\begin{figure}[t!]
    \centering
    \setlength{\abovecaptionskip}{0pt}
    \setlength{\belowcaptionskip}{0pt}
    \begin{subfigure}{0.4\linewidth}
        \includegraphics[width=\linewidth]{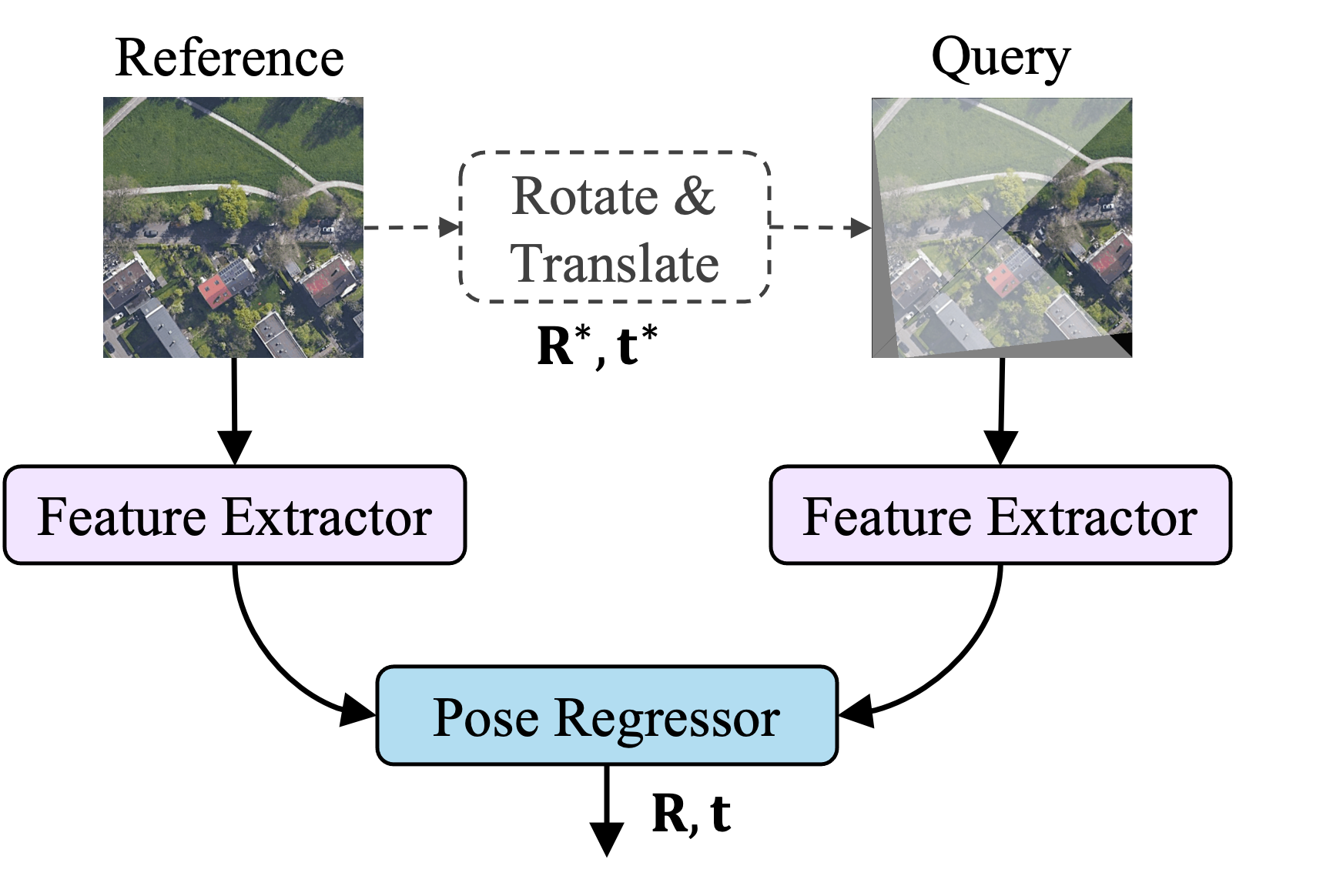}
        \caption{\scriptsize Training}
    \label{subfig:training}
    \end{subfigure}
    \hspace{1em}
    \begin{subfigure}{0.4\linewidth}
        \includegraphics[width=\linewidth]{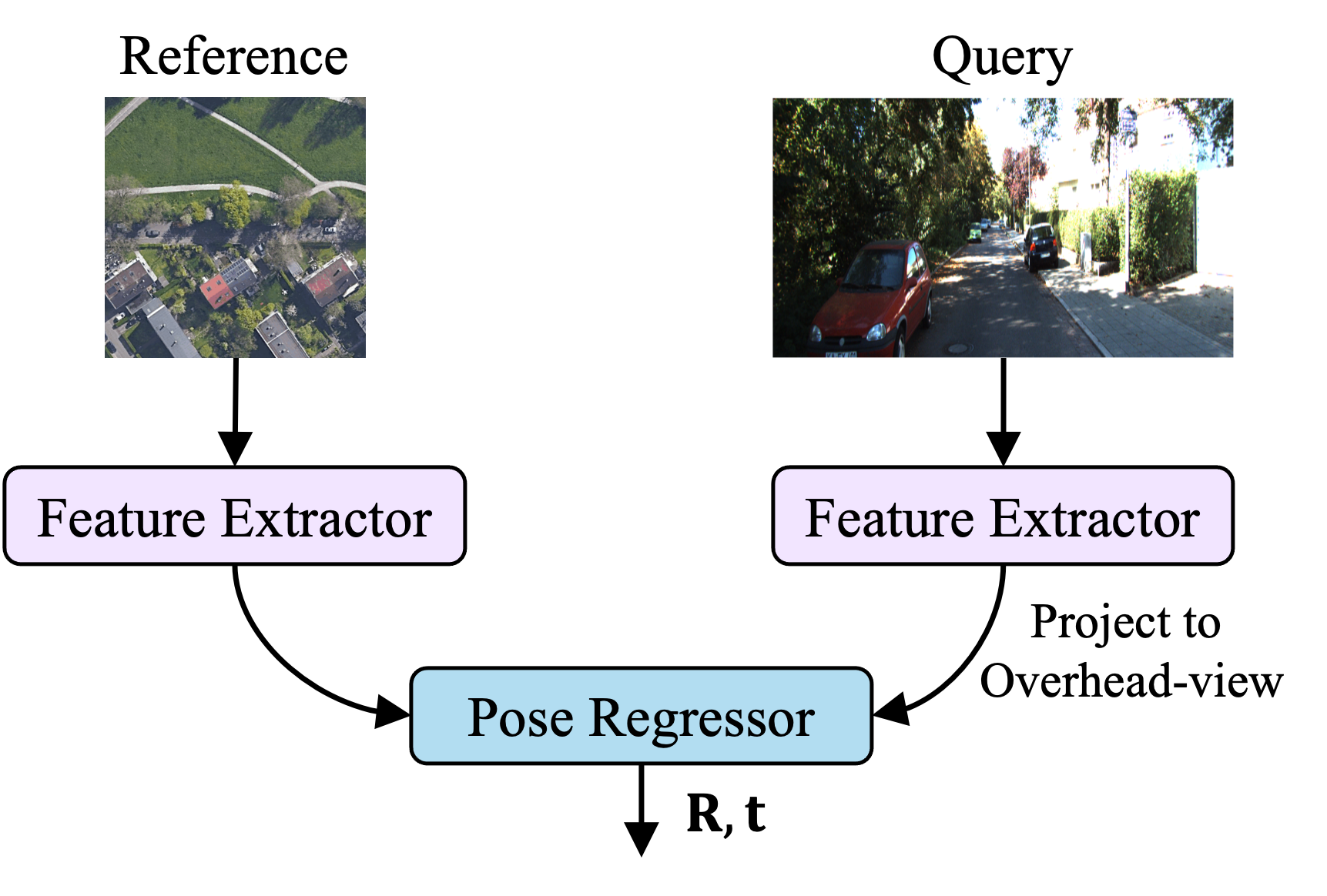}
        \caption{\scriptsize Testing}
    \label{subfig:testing}
    \end{subfigure}
    \caption{\scriptsize Self-supervised rotation estimator.
    }
    \label{fig:frame_rot}
\end{figure}

\textbf{Increasing localization accuracy.}
Recently, researchers have started to investigate how to increase the accuracy of a ground camera's pose by ground-to-satellite image matching once the camera's coarse rotation and orientation have been determined.
Towards this purpose, network regression~\cite{zhu2021vigor}, pose optimization~\cite{shi2022beyond}, similarity matching~\cite{xia2022visual,lentsch2023slicematch,xia2023convolutional,fervers2022uncertainty,shi2023boosting,shi2022accurate,sarlin2023orienternet,zhang2024increasing}, correspondence learning~\cite{song2024learning}, and iterative tomography estimation~\cite{wang2024fine} have been explored. 
Nonetheless, all these works need sub-meter and sub-degree pose labels for ground images in the training data to train their networks. In this paper, we propose a strategy that estimates the relative rotation and translation between a ground and a satellite image when such accurate labels are unavailable in the training data.


\subsection{Self/weakly-supervised learning}
Self- and weakly-supervised learning has been widely explored in image classification~\cite{Zhai_2019_ICCV}, object detection~\cite{dang2023study}, semantic segmentation~\cite{wang2022fully}, image inpainting~\cite{pathak2016context}, point cloud registration~\cite{liu2023self}, human/hand/object pose estimation~\cite{bouazizi2021self,spurr2021self,gharaee2023self}, the intersection between vision and natural language processing~\cite{radford2021learning} \etc. 
Many of them also exploit contrastive learning as supervision signals. 
Tang~\etal~\cite{tang2021self} proposed a self-supervised strategy for localizing outdoor range sensors (\eg, Lidar, Radar). 
However, little attention has been paid to camera pose estimation with self or weak supervision. 
Our concurrent work~\cite{xia2024adapting} also solves the problem of weakly supervised ground-to-satellite camera localization, but it highlights how to adapt a model trained with GT supervision from one area to another area where only coarse supervision is available. 
In contrast, this paper does not assume any GT data from any area. 

\section{Method}

Given a coarse location and orientation of a ground camera, our goal is to refine this pose through ground-to-satellite image registration. In contrast to previous works that assume a large training dataset with highly accurate pose labels for ground images, we propose a weakly supervised learning strategy that does not require such labels. 
Our approach first estimates the relative rotation by network regression and then computes the translation by similarity matching.



\subsection{Rotation Estimation by Self-supervision}
\label{sec:R}
We draw inspiration from Spatial Transformer Networks \cite{jaderberg2015spatial} to regress relative rotations between cross-view images by a network. The network takes query and reference images as input and outputs their relative pose, as illustrated in Fig.~\ref{fig:frame_rot}.  

\textbf{Training.} To supervise the network, we generate some ``satellite-and-satellite'' image pairs with GT relative poses. 
This is done by rotating and translating a satellite image using a randomly generated pose,  $\mathbf{R}^*(\theta), \mathbf{t}^*$,  where the maximum magnitude of the rotation {\color{black} angle} $\theta$ and translation is based on the statistical error of the ground camera's coarse pose information that we aim to refine during deployment. 
For example, when the coarse location is achieved by ground-to-satellite image retrieval, the minimum statistical location error (assuming the top-1 retrieval recall rate is 100\%) is the half distance between centers of two nearest satellite images, \ie, the half grid size when sampling satellite images for constructing the database~\cite{zhu2021vigor,shi2022accurate,shi2023cvlnet}. 
Furthermore, we apply a mask on the transformed satellite image and extract a triangle region corresponding to the ground camera's horizontal FoV, as shown at the top of Fig.~\ref{subfig:training}.
This is intended to mimic a synthesized overhead view image from a ground-view image. 
We refer to the transformed and masked satellite image as the query and the original one as the reference. 
We then train the network to estimate the relative pose between the two satellite images. 
The training objective is:

{\color{black}

\begin{equation}
    \mathcal{L}_1 =\left | \theta - \theta^* \right | + \left | t_x - t_x^* \right |  + \left | t_y - t_y^* \right |,
\end{equation}
where $\theta, t_x$ and $t_y$ denote the network predictions, $\theta^*, t_x^*$ and $t_y^*$ indicate the corresponding GT, $\theta$ is the 1-DoF inplane rotation, \ie, the yaw angle, $t_x$ and $t_y$ represent the 2-DoF translation,} and $\left| \cdot \right|$ denote the $L_1$ norm. 

\textbf{Evaluation.} After the network is trained, we substitute the query satellite image with the query ground images during inference for the ground camera's pose estimation, as shown in Fig.~\ref{subfig:testing}.
We found the feature extractors for the satellite images and ground images captured by a perspective pin-hole camera are sharable. 
A potential explanation might be that the projection geometry of them is similar. They both map straight lines in the real world to straight lines in images.
We demonstrate this by experiments in the main paper and supplementary material. 
As the pose regressor accepts query and reference feature maps in the overhead view, we follow previous works~\cite{shi2023boosting,shi2023cvlnet} to project ground-view feature maps to overhead view by exploiting the ground plane Homography. 
This projection method does not require additional trainable parameters and thus allows our model trained between satellite-and-satellite images deployable between satellite-and-ground images. 
Furthermore, since the ground-plane Homohraphy establishes correct correspondences for scene contents on the ground plane between the two views, which occupies a large portion of the field-of-view of the ground camera in autonomous driving scenarios,  
our pose regressor handles well when the query feature map is the projected one from the ground view.  

Our pose regressor shares the same pros and cons as the neural pose optimizer in Shi et al.~\cite{shi2023boosting}. 
It leverages the sensitiveness of general neural networks (without specific designs) to rotations on input signals~\cite{worrall2017harmonic,Lee_2023_CVPR}, and thus achieves promising results on rotation estimation. However, the translation estimation performance is poor due to the aggregation layers, \eg, max-pooling, inside the pose regressor, making it insensitive to slight translations on input signals. 
To recover accurate relative translations, we leverage the translational equivariance of spatial correlation (a.k.a. convolution), which is described below.   

\begin{figure}[t!]
    \centering
    \setlength{\belowcaptionskip}{0pt}
    \includegraphics[width=\linewidth]{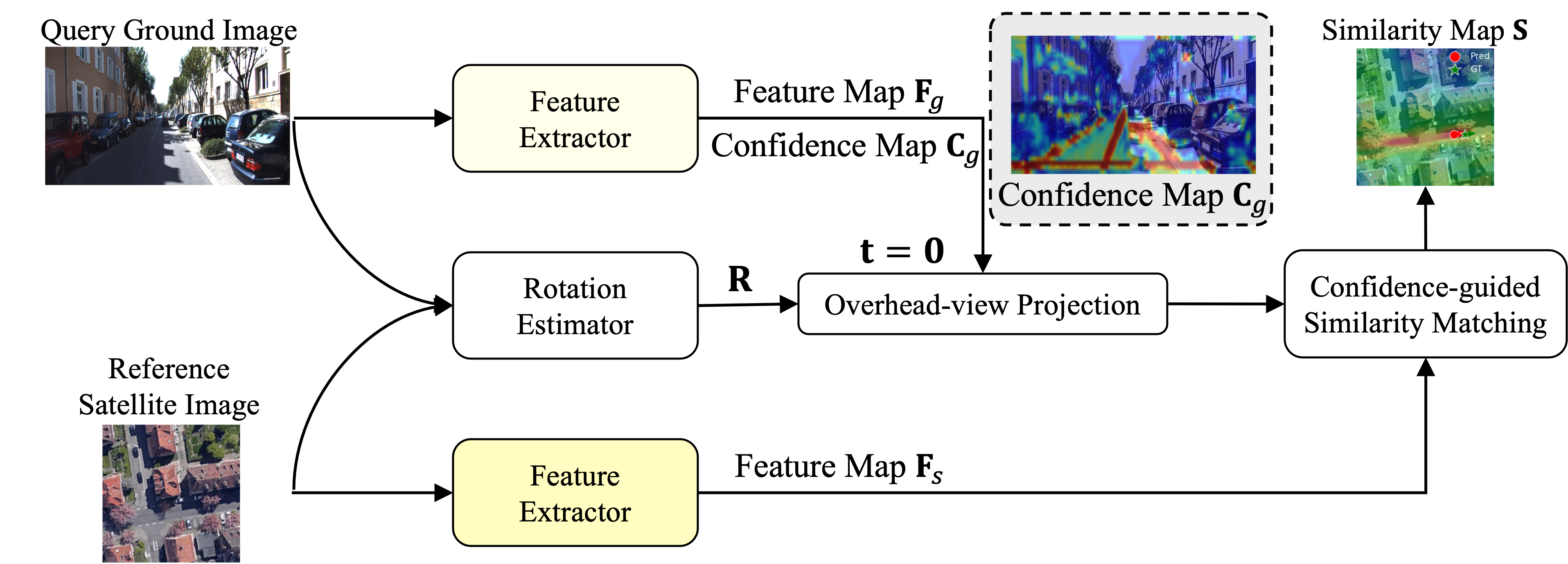}
    \caption{\scriptsize Translation estimation framework. During training, the weights of the rotation estimator are fixed, and we only train the two-branch feature extractors. 
    Other uncolored blocks indicate no trainable parameters are involved. 
    }
    \label{fig:framework}
\end{figure}

\subsection{Translation Estimation by Deep Metric Learning}
\label{sec:T}
As shown in Fig.~\ref{fig:framework}, a two-branch convolutional network is first applied to the ground and satellite image pair. 
Each branch is a U-Net architecture and extracts multi-level representations of the original images.

For the ground branch, not only the feature representation $\mathbf{F}_{g}\in \mathbb{R}^{H_g \times W_g \times C}$ but also a confidence map $\mathbf{C}_{g} \in \mathbb{R}^{H_g \times W_g \times 1}$ is extracted. 
The confidence map is an additional channel of the feature extractor output followed by a sigmoid layer. It indicates whether features at corresponding spatial pixel positions are trustworthy.
For example, dynamic objects (\eg, cars) in the images are detrimental to localization performance, while the static road structures are essential features. 
The higher the confidence, the more reliable the corresponding features. 
We should note that no explicit supervision is applied to the confidence map. Instead, it is encoded in the cross-view similarity-matching process and learned statistically from the similarity-matching training objective. 

We only extract feature representations $\mathbf{F}_{s}\in \mathbb{R}^{H_s \times W_s \times C}$ for the satellite branch with no confidence map because we empirically found learning confidence maps for satellite images impairs the performance. The reasons hypothesized are twofold: \textbf{(i)} dynamic objects are fewer on the satellite images, and they occupy a relatively smaller region on the satellite image compared to ground-view images; thus, they have a lower impact on localization performance; \textbf{(ii)} neither the camera poses, nor the confidence maps, have explicit supervision, thus learning a confidence map for satellite images increases the network training difficulty.

We leverage the trained rotation estimator in Sec.~\ref{sec:R} to estimate the relative rotation between the ground and satellite images. 
Then, the ground-view features and confidence maps are projected to the overhead view according to the estimated rotation 
$\mathbf{R}$ and zesro translation $\mathbf{t} = \mathbf{0}$, exploiting the ground plane Homography. 
To maintain claritys, we utilize the original symbols $\mathbf{F}_g$ and $\mathbf{C}_g$ to represent the projected ground features and confidence maps.

\textbf{Confidence-guided similarity matching.}
Given that the rotation of the projected overhead-view feature map has been aligned with the observed satellite feature map, the only remaining disparity between them is a translation difference. We explore the normalized spatial correlation to compute this translation difference. Specifically, the projected overhead-view feature map is used as a sliding window, and its normalized inner product with the reference satellite feature map is computed when aligned at varying locations. 
The mathematical representation, 
taking into account ground-view confidence maps, is as follows:
\begin{equation}
    \mathbf{S}(u,v) = ({\mathbf{F}_s} * \widehat{\mathbf{F}_g})(u, v) = \frac{\sum_i \sum_j {\mathbf{F}_s} (u + i, v+j) \widehat{\mathbf{F}_g}(i, j) }{\sqrt{\sum_i \sum_j {\mathbf{F}_s}^2 (u + i, v+j)}\sqrt{\sum_i \sum_j\widehat{\mathbf{F}_g}^2(i, j) }},
\end{equation}
where $\widehat{\mathbf{F}_g} = \mathbf{C}_g\mathbf{F}_g$ is to highlight important features while suppressing non-reliable features for localization.
The result $\mathbf{S}(u,v)$ is actually the cosine similarity between the two view features when aligned at each location $(u,v)$.
Thus, the pixel coordinate corresponding to the maximum similarity, $\arg\max_{(u, v)} \mathbf{S}$,  indicates the most likely ground camera's location. This similarity map between the synthesized overhead-view feature map and the reference satellite feature map can also be regarded as the location probability map of the ground camera. 

\textbf{Supervision.} 
We apply deep metric learning for network supervision. For a query ground image, we compute a similarity map based on the possible locations between it and its positive and negative satellite images, denoted as  $\mathbf{S}_{\text{pos}}$ and $\mathbf{S}_{\text{neg}}$, respectively. We maximize the maximum similarity in $\mathbf{S}_{\text{pos}}$ while minimizing the maximum similarity in $\mathbf{S}_{\text{neg}}$:

\begin{equation}
    \mathcal{L}_2 = \sum_l log(1 + e^{\alpha(\max\mathbf{S}_{\text{neg}}-\max\mathbf{S}_{\text{pos}})}), 
    \label{eq:contrastive}
\end{equation}
where $\alpha$ controls the convergence speed and is set to 10. 

When the positive image pairs in the training dataset are created in the same way as image pairs during inference, for example, by the same ground-to-satellite retrieval model or the same noisy GPS sensor, the error of location labels in the training set is the same as the error of locations that we aim to refine during deployment. 
In this case, we only employ Eq.~\eqref{eq:contrastive} for network training. In another scenario where relatively more accurate location labels for ground images in the training data are available than the poses we aim to refine during employment, we introduce an additional training objective to incorporate this signal: 
\begin{equation}
    \mathcal{L}_3 = \sum_l \left | \max(\mathbf{S}_{\text{pos}}) -  \max(\mathbf{S}_{\text{pos}} [u^* - \frac{d}{\gamma} :u^* + \frac{d}{\gamma} , v^* - \frac{d}{\gamma}  :v^* + \frac{d}{\gamma}])\right |,
    \label{eq:weak}
\end{equation}
where $(u^*, v^*)$ indicates the location label provided by the training data and has an error of up to $d$ meters, $\gamma$ denotes the ground resolution of the similarity map in terms of meters per pixel. This training objective forces the global maximum in the similarity map to equal a local maximum, with the local region centered at the location label with a radius of $d$ meters. 

The whole training objective is: 
\begin{equation}
    \mathcal{L} = \mathcal{L}_2 + \lambda\mathcal{L}_3,
\end{equation}
where $\lambda=0$ indicates such relatively accurate pose labels are unavailable in the training set, while $\lambda=1$ suggests such labels are available. In our experiments, we set $d=5$ meters.

\textbf{Overall Evaluation. }
After training the networks, the overall evaluation goes through the framework shown in Fig.~\ref{fig:framework}. 
The input is a query image and its positive satellite image, and the output is an estimated relative rotation between the two images and a location probability map of this query image with respect to this satellite image. The location corresponding to the maximum probability/similarity value is deemed as the query camera location.

\section{Experiments}

\quad \ \textbf{Network Architectures.} 
A UNet-based architecture with a pre-trained VGG16 as the encoder is adopted for feature extraction. The decoder of the UNet is randomly initialized. 
We empirically found that the feature extractor of satellite images is shareable with ground images captured by a pin-hole camera while not shareable with ground panoramas. 
This might be because both satellite images and ground images captured by a pin-hole camera map straight lines in the real world to straight lines on images, while panoramas map straight lines in the real world to curves on images. 
While for different purposes, \ie, rotation and translation estimation, we found non-shareable feature extractors between the two stages help to achieve the best performance. 
We present detailed analysis and experimental demonstrations in the supplementary material. 
The pose regressor in Fig.~\ref{fig:frame_rot} is constructed by two swin transformer layers~\cite{liu2021swin} followed by two FC layers. This is the same as the neural optimizer architecture in \cite{shi2023boosting}.

\textbf{Dataset.} Our experiments are conducted on a well-known autonomous driving dataset, KITTI~\cite{Geiger2013IJRR}, and a cross-view localization dataset, VIGOR~\cite{zhu2021vigor}. 

For the {KITTI} dataset, ground images were captured by a forward-facing pin-hole camera with a limited FoV. The cross-view KITTI dataset includes one training set and two testing sets. Images from Test-1 are from the same region as the training set, while images from Test-2 are from a different region. 
The location search range for this dataset is around $56 \times 56$ m$^2$, and the orientation noise is $20^\circ$, which follows the official setting as in \cite{shi2022beyond}. 
The performance evaluation on different coarse pose errors is presented in the supplementary material. 

The {VIGOR} dataset contains ground and satellite images from four cities in the US: Chicago, New York, San Francisco, and Seattle. It is divided into same-area and cross-area splits. The same-area split indicates the training and testing images are from the same region (both from the four cities), while the cross-area split adopts images from two cities for training and images from the other two cities for testing. 
In the original dataset, each ground image has a positive satellite image and several semi-positive satellite images, depending on whether this ground image is within the center $\frac{1}{4}$ region of the satellite image. 
We follow \cite{lentsch2023slicematch} and use only the positive satellite images.

Since the feature extractors trained for satellite images (Fig.~\ref{subfig:training}) are not applicable to panoramas due to differences in imaging modality, we only evaluate the translation estimation performance on the VIGOR dataset with known and unknown (360$\circ$ ambiguity) orientations, respectively. 
The 3-DoF joint location and orientation pose estimation is performed on the KITTI dataset.

\textbf{Evaluation Metrics.} 
Following the approach of \cite{shi2022beyond}, we decompose the translation along the lateral (orthogonal to driving direction) and longitudinal (along the driving direction) directions for evaluation on the KITTI dataset. Specifically, for a query image, we consider it to be successfully localized along a direction if its estimated location along that direction is within $d$ meters of its ground truth (GT) location. Similarly, we consider the rotation estimation correct if the estimated rotation is within $\theta^\circ$ of the GT rotation. We record the percentage of successfully localized images along each direction and the percentage of images with correct rotation estimation.

The VIGOR dataset does not provide information on the driving direction of the camera. Therefore, we cannot decompose the translation to be along lateral and longitudinal directions.
On this dataset, we report the median and mean errors of comparison algorithms, following the approach of \cite{xia2022visual} and \cite{lentsch2023slicematch}.

\textbf{Data augmentation and implementation Details. } 
For training the rotation estimator, we use the positive satellite image for each ground image and randomly transform each image once to create training input pairs. 
We adopt feature size as $\frac{1}{4}$ of the original image size for both rotation and translation estimation. The original image size is not used because of its large memory consumption in the spatial correlation process. 
We use a batch size of $B=8$ to train the network. 
In the translation estimation training, each ground image has one matching satellite image and $B-1$ non-matching satellite images within each batch. 
Our experiments are conducted on an RTX 3090 GPU. The network is trained for $3$ epochs for both stages on the KITTI dataset and $10$ epochs on the VIGOR dataset.
For the ground images in the KITTI dataset, we use a resolution of $256 \times 1024$. For the VIGOR dataset, we use a resolution of $320\times 640$. The satellite image resolution is $512\times512$ for all datasets. The ground resolution of satellite images is $0.2$ meters per pixel in the KITTI dataset, and $0.111$, $0.113$, $0.118$, and $0.101$ meters per pixel for the cities Chicago, New York, San Francisco, and Seattle in the VIGOR dataset, respectively.
The source code of this paper is available at \url{https://github.com/YujiaoShi/G2SWeakly.git}.

\begin{figure}[t!]
\centering
\setlength{\abovecaptionskip}{0pt}
\setlength{\belowcaptionskip}{0pt}
\includegraphics[width=\linewidth]{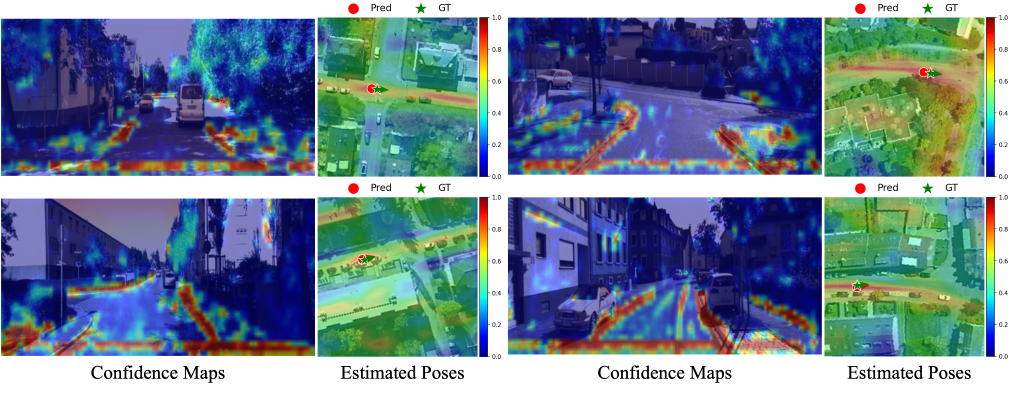}
    \caption{\scriptsize Visualization of learned confidence maps for ground images and estimated poses by our method. 
    }
    \label{fig:visualization}
\end{figure}

\begin{table}[t]
\centering
\setlength{\abovecaptionskip}{0pt}
\setlength{\belowcaptionskip}{0pt}
\setlength{\tabcolsep}{0pt}
\scriptsize 
\caption{\scriptsize Ablation study on the cross-view KITTI dataset. 
We report the percentage of query images whose locations are restricted to be within $d$ meters of their GT locations along lateral/longitudinal directions and whose orientations are restricted to be within $\theta^\circ$ of their GT orientation, respectively.  
}
\begin{tabular}{c|c|c|cH|ccH|ccH|ccH|ccH|ccH|ccH}
\toprule
\multirow{3}{*}{Query} & \multirow{3}{*}{$\mathbf{t}$ Est.} & \multirow{3}{*}{Conf} & \multirow{3}{*}{$\lambda$} & \multirow{3}{*}{Label Error} & \multicolumn{9}{c|}{Test-1 (Same-area)}                                                         & \multicolumn{9}{c}{Test-2 (Cross-area)}                                                         \\
                             &                         &                       &                            &                              & \multicolumn{3}{c|}{Lateral}                          & \multicolumn{3}{c|}{Longitudinal}                         & \multicolumn{3}{c|}{Azimuth}                      & \multicolumn{3}{c|}{Lateral}                          & \multicolumn{3}{c|}{Longitudinal}                         & \multicolumn{3}{c}{Azimuth}                        \\
                             &                         &                       &                            &                               & $d=1\uparrow$          & $d=3\uparrow$          & $d=5\uparrow$          & $d=1\uparrow$         & $d=3\uparrow$          & $d=5\uparrow$          & $\theta=1\uparrow$     & $\theta=3\uparrow$     & $\theta=5$     & $d=1\uparrow$          & $d=3\uparrow$          & $d=5$          & $d=1\uparrow$         & $d=3\uparrow$          & $d=5$          & $\theta=1\uparrow$     & $\theta=3\uparrow$     & $\theta=5$     \\\midrule
Sat        & Reg.                     & --                    & --                         & --                          & 4.88	&15.13	&25.18	&4.80	&15.64	&25.36	&99.89	&100.00	&100.00	&5.06	&15.46	&25.60	&5.25	&15.79	&25.56	&99.95	 & 100.00 & 100.00 \\
Grd        & Reg.                    & --                    & --                         & --                           & 4.88   & 15.11  & 25.15 & 4.74   & 15.66  & 25.34 & 99.66  & 100.00 & 100.00 & 5.04   & 15.46  & 25.58 & 5.29   & 15.83  & 25.56 & 99.99  & 100.00 & 100.00 \\
Grd        & Corr.                    & N                     & 0                          & --                      & 45.80	&78.27	&84.73	&6.18	& 16.67	& 26.03	& 99.66	& 100.00	& 100.00 & 45.11 & 73.04	&82.87	&6.13	&18.30	&26.57	&99.99  & 100.00 & 100.00 \\
Grd        & Corr.                   & Y                     & 0                          & --                       & 59.58  &85.74	&90.38	&11.37	&31.94	&44.42	&99.66	&100.00	&100.00	&62.73	&86.53	&90.61	&9.98	&29.67	&41.29	&99.99	&100.00	&100.00\\
Grd        & Corr.                   & Y                     & 1                          & --                        &\textbf{66.07}	&\textbf{94.22}	 &\textbf{97.93}	&\textbf{16.51}	&\textbf{49.96}	&\textbf{67.96}	&\textbf{99.66}	&\textbf{100.00}	&\textbf{100.00}	&\textbf{64.74}	&\textbf{86.18}	&\textbf{89.27}	&\textbf{11.81}	&\textbf{34.77}	&\textbf{46.83}	&\textbf{99.99}	&\textbf{100.00}	&\textbf{100.00}\\
\bottomrule 
\end{tabular}
\label{tab:ablation}
\end{table}

\subsection{Ablation Study}


Below, we demonstrate the necessity and effectiveness of each proposed component. The results are presented in Tab.~\ref{tab:ablation}.

\textbf{(i)} The first row shows the performance of our pose regressor (Reg.) in Fig.~\ref{fig:frame_rot} with the satellite images as queries (note: not ground images). It can be seen that the estimated rotation of almost all the queries has been restricted to within $1^\circ$ of its GT rotation, while the translation estimation performance is poor, even though there is no domain gap between reference and query images. This supports our intuition that a neural network-based regressor's ability to provide accurate translation estimation is limited. 

\textbf{(ii)} Then, we verify the generalization ability of the pose regressor on real ground images. The results in the second row show that the rotation estimation accuracy for real ground images is very close to that for satellite images, demonstrating the effectiveness of our self-supervision training strategy by satellite-and-satellite image pairs.

\textbf{(iii)} The third row presents the performance of our method by using spatial correlation for translation estimation and pose regressor for rotation estimation. 
The performance of translation estimation is significantly boosted.

\textbf{(iv)} In what follows, we encode the confidence in the spatial correlation process. The results in the fourth row demonstrate the effectiveness of the confidence-guided similarity matching. 
Fig.~\ref{fig:visualization} provides visualizations of the learned confidence maps and the probability maps of query cameras' location with respect to their matching satellite images. It can be seen that the learned confidence maps are able to ignore dynamic objects and highlight reliable features (e.g., lane lines, road edges).

\textbf{(v)} Finally, when relatively accurate (but still noisy) pose labels are available in the training dataset, \ie, the pose error is up to $d=5$ meters, we set $\lambda=1$ and use Eq.~\ref{eq:weak} to incorporate this information in training. The last row of Tab.~\ref{tab:ablation} demonstrates that it successfully boosts the network learning process and improves the performance on both same-area and cross-area evaluation.

\begin{table}[t!]
\centering
\setlength{\abovecaptionskip}{0pt}
\setlength{\belowcaptionskip}{0pt}
\setlength{\tabcolsep}{0pt}
\scriptsize 
\caption{\scriptsize Comparison between different overview synthesis methods on KITTI.
}
\begin{tabular}{c|c|ccH|ccH|ccH|ccH|ccH|ccH}
\toprule
 \multicolumn{1}{c|}{ \multirow{3}{*}{\begin{tabular}[c]{@{}c@{}}Overhead-view\\ Feat. Syn.\end{tabular}}}   & \multicolumn{1}{c|}{ \multirow{3}{*}{$\lambda$}}                               & \multicolumn{9}{c|}{Test-1 (Same-area)}                                                         & \multicolumn{9}{c}{Test-2 (Cross-area)}                                                         \\
                            \multicolumn{1}{c|}{}    &                       & \multicolumn{3}{c|}{Lateral}                          & \multicolumn{3}{c|}{Longitudinal}                         & \multicolumn{3}{c|}{Azimuth}                      & \multicolumn{3}{c|}{Lateral}                          & \multicolumn{3}{c|}{Longitudinal}                         & \multicolumn{3}{c}{Azimuth}                        \\
                             \multicolumn{1}{c|}{}  &          & $d=1\uparrow$          & $d=3\uparrow$          & $d=5$          & $d=1\uparrow$         & $d=3\uparrow$          & $d=5$          & $\theta=1\uparrow$     & $\theta=3\uparrow$     & $\theta=5$     & $d=1\uparrow$          & $d=3\uparrow$          & $d=5$          & $d=1\uparrow$         & $d=3\uparrow$          & $d=5$          & $\theta=1\uparrow$     & $\theta=3\uparrow$     & $\theta=5$     \\\midrule
Geo. Trans. \cite{shi2023boosting}  &\multirow{2}{*}{0}  & 48.74 & 83.14 & 89.77 & 8.27  & 24.20 & 34.53 & 99.66 & 100.00 & 100.00 & 52.69 & 79.99 & 84.99 & 8.46  & 24.00 & 34.08 & 99.99 & 100.00 & 100.00 \\
\textbf{Ours}                 &                    & 59.58 & 85.74 & 90.38 & 11.37 & 31.94 & 44.42 & 99.66 & 100.00 & 100.00 & 62.73 & 86.53 & 90.61 & 9.98  & 29.67 & 41.29 & 99.99 & 100.00 & 100.00 \\ \midrule
Geo. Trans. \cite{shi2023boosting}  & \multirow{2}{*}{1} &62.02 & 92.13 & 97.69 & 16.01 & 46.54 & 64.59 & 99.66 & 100.00 & 100.00 & 62.97 & 87.54 & 91.04 & 11.36 & 33.70 & 45.80 & 99.99 & 100.00 & 100.00 \\
\textbf{Ours}                  &                    &66.07 & 94.22 & 97.93 & 16.51 & 49.96 & 67.96 & 99.66 & 100.00 & 100.00 & 64.74 & 86.18 & 89.27 & 11.81 & 34.77 & 46.83 & 99.99 & 100.00 & 100.00 \\
\bottomrule   
\end{tabular}
\label{tab:projGeoCrossAttn}
\end{table}

\subsection{Different overhead-view projections}

In this section, we ablate the performance of our method with different overhead-view projections. 
Among existing works, OrienterNet~\cite{sarlin2023orienternet} and Shi et al.~\cite{shi2023boosting} both propose geometry-guided cross-view transformers (Geo. Trans.) for overhead-view feature synthesis. 
Shi et al.~\cite{shi2023boosting} achieves slightly better performance, as illustrated in Tab.~\ref{tab:kitti}. 
Thus, we evaluate the performance of the Geo. Trans. from Shi et al.~\cite{shi2023boosting} in our framework. 
From the comparison results in Tab.~\ref{tab:projGeoCrossAttn}, it can be seen the performance of Geo. Trans is inferior to the simple ground plane Homography-based overhead-view projection under the weakly-supervised setting. 
We hypothesize two main reasons for this: 

{\textbf{(i)} Smaller batch size. } In this paper, we leverage deep metric learning for network supervision. 
It is well-recognized that a small batch size in metric learning affects performance negatively. 
SimpleBEV \cite{harley2023simple} also confirms that a larger batch size improves the overall performance more significantly compared to different overhead-view projection methods.  
With our simple Homography-based projection method, we achieved a batch size of 8 on an RTX 3090 GPU, whereas the batch size is 4 when using the Geo. Trans. due to its complexity.

{\textbf{(ii)} Weak supervision. } 
The previous geometry-guided cross-view transformers are usually constructed with complex designs to make the learned overhead view feature representations invariant to parallel, occlusion, and appearance changes. 
This works well when the supervision signal is strong, \ie, GT pose labels are available. 
However, in our scenario, the pose labels are noisy, and only weak supervision is applied. 
This limits the learning ability of geometry-guided cross-view transformers. 
Instead, our method explores the explicit ground plane Homography for the overhead-view feature projection. 
It requires no additional trainable parameters and reduces the learning burden of deep networks. 
The ground plane Homography handles the co-visible regions for the overhead-view feature synthesis, \ie, scenes on the ground plane. 
For not co-visible regions, our confidence map lowers their weights in the similarity matching process for camera localization. 

We should note that we do not want to claim our solution is optimal for overhead-view synthesis in the weakly-supervised ground-to-satellite camera localization task. 
We believe there should be more advanced approaches that make the best balance between capability, complexity, and the easiness of training under the weak supervision scenario. 
The main purpose of this paper is to introduce training strategies for pose estimation under the weak supervision scenario. 
In order to not dilute our contributions, we leave the exploration of a better overhead-view feature synthesis module as our future work.

\begin{figure}[t!]
    \centering
    \setlength{\abovecaptionskip}{0pt}
    \setlength{\belowcaptionskip}{0pt}
    \includegraphics[width=0.85\linewidth]{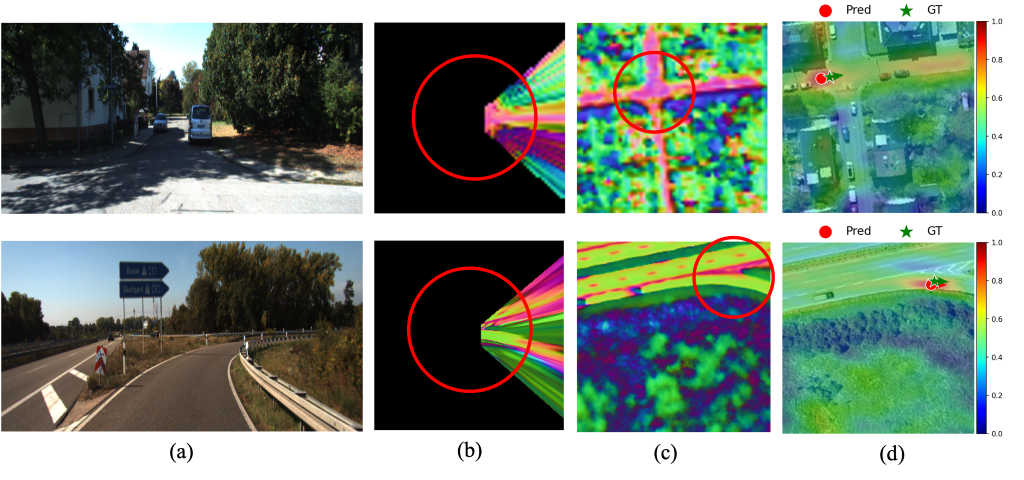}
    \caption{(a) Ground image; (b) Synthesized overhead view feature map from the ground image;
    (c) Satellite image feature map; (d) Estimated location probability map and camera orientation indicated by the red arrow. 
    }
    \label{fig:feature_map}
\end{figure}

\textbf{Visualization of learned overhead-view features. }
We visualize the synthesized overhead view feature maps from ground images in Fig.~~\ref{fig:feature_map} (b). The center of the synthesized overhead view feature map corresponds to the ground camera location, and the right direction corresponds to the heading direction (z-axis) of the ground camera. The corresponding satellite image features are presented in Fig.~\ref{fig:feature_map} (c). 
Using our framework, we register the query feature map Fig.~\ref{fig:feature_map} (b) to the reference feature map Fig.~\ref{fig:feature_map} (c). The resulting camera orientation and location probability maps are shown in Fig.~\ref{fig:feature_map} (d). 
To reduce ambiguity, we only consider the scene content within 20m of the query camera location for localization, marked as red circles in Fig.~\ref{fig:feature_map}.
It can be seen that the scene contents on the ground plane are recovered faithfully. 
A visual illustration of this spatial correlation process is provided in the supplementary material.

\subsection{Comparison with the state-of-the-art}

In this section, we compare the performance of our method with the recent state-of-the-art, including DSM~\cite{shi2020looking}, CVR~\cite{zhu2021vigor}, \cite{shi2022beyond}, MCC~\cite{xia2022visual}, SliceMatch~\cite{lentsch2023slicematch}, {\color{black} OrienterNet~\cite{sarlin2023orienternet}}, Xia~\etal~\cite{xia2023convolutional}, Shi~\etal~\cite{shi2023boosting}, and Song~\etal~\cite{song2024learning}. 
All these state-of-the-art algorithms adopt full supervision and rely on accurate pose labels for their network, and their results are either taken from their original papers or re-evaluated using the author-released models. 

\begin{table}[t!]
\centering
\setlength{\abovecaptionskip}{0pt}
\setlength{\belowcaptionskip}{0pt}
\setlength{\tabcolsep}{0pt}
\scriptsize 
\caption{\scriptsize Comparison with the state-of-the-art on KITTI. $*$ indicates full supervision is adopted. 
}
\begin{tabular}{HccHH|ccH|ccH|ccH|ccH|ccH|ccH}
\toprule
\multirow{3}{*}{Supervision} & \multicolumn{4}{c|}{\multirow{3}{*}{Algorithms}}  & \multicolumn{9}{c|}{Test-1 (Same-area)}                                                         & \multicolumn{9}{c}{\textbf{Test-2 (Cross-area)}}                                                         \\
                             &                         &                       &                            &                              & \multicolumn{3}{c|}{Lateral}                          & \multicolumn{3}{c|}{Longitudinal}                         & \multicolumn{3}{c|}{Azimuth}                      & \multicolumn{3}{c|}{Lateral}                          & \multicolumn{3}{c|}{Longitudinal}                         & \multicolumn{3}{c}{Azimuth}                        \\
                             &                         &                       &                            &                               & $d=1\uparrow$          & $d=3\uparrow$          & $d=5$          & $d=1\uparrow$         & $d=3\uparrow$          & $d=5$          & $\theta=1\uparrow$     & $\theta=3\uparrow$     & $\theta=5$     & $d=1\uparrow$          & $d=3\uparrow$          & $d=5$          & $d=1\uparrow$         & $d=3\uparrow$          & $d=5$          & $\theta=1\uparrow$     & $\theta=3\uparrow$     & $\theta=5$     \\\midrule
\multirow{5}{*}{Fully}       & \multicolumn{4}{c|}{DSM~\cite{shi2020looking}$^*$}                                                                              &10.12          & 30.67          & 48.24          & 4.08          & 12.01          & 20.14          & 3.58           & 13.81          & 24.44          & 10.77          & 31.37          & 48.24          & 3.87          & 11.73          & 19.50          & 3.53           & 14.09          & 23.95        \\
                            & \multicolumn{4}{c|}{CVR~\cite{zhu2021vigor}$^*$}                                                                              & 18.61          & 49.06          & 69.79          & 4.29          & 13.01          & 21.47          & -              & -              & -              & 17.38          & 48.20          & 70.79          & 4.07          & 12.52          & 20.14          & -              & -              & -  \\
                             & \multicolumn{4}{c|}{Shi and Li\cite{shi2022beyond}$^*$}                                                                              & 35.54  & 70.77  & 80.36 & 5.22   & 15.88  & 26.13 & 19.64  & 51.76  & 71.72  & 27.82  & 59.79  & 72.89 & 5.75   & 16.36  & 26.48 & 18.42  & 49.72  & 71.00  \\
                             & \multicolumn{4}{c|}{SliceMatch~\cite{lentsch2023slicematch}$^*$}                                                                              & 49.09  &  91.76   & 98.52 & 14.19  &  49.99  & 57.35 & 13.41  &  42.62  & 64.17  & 32.43  & 78.98   & 86.44 & 8.30   & 24.48  & 35.77 & 46.82  & 46.82   & 46.82  \\ 

                            &  \multicolumn{4}{c|}{{\color{black}OrienterNet\cite{sarlin2023orienternet}$^*$}} & - & - & - & - & - & - & - & - & - & {\color{black}51.26}     & {\color{black}84.77} & -     &  {\color{black}22.39}     &{\color{black}46.79}   & - &   {\color{black}20.41}   & {\color{black}52.24} & - \\ 
                            & \multicolumn{4}{c|}{Xia et al.\cite{xia2023convolutional}}$^*$    & \textbf{97.35} &\textbf{98.65} &99.71 &77.13 &\textbf{96.08} &97.16 & 77.39 &99.47 &99.95 &44.06 &81.72 &90.23 &23.08 &52.85 &64.31& 57.72 &92.34 &96.19 \\ 
                             & \multicolumn{4}{c|}{Shi et al. \cite{shi2023boosting}$^*$}       & {76.44}	&{96.34}	&{98.89}	&{23.54}	&{50.57}	&{62.18}	&{99.10}	&\textbf{100.00}	&\textbf{100.00} &{57.72}	&\textbf{86.77}	&{91.16}	&{14.15}	&{34.59}	&{45.00}	&{98.98}	&\textbf{100.00}	&\textbf{100.00}\\
                             & \multicolumn{4}{c|}{Song et al. \cite{song2024learning}$^*$}       & {95.47}	&--	&{99.79}	&\textbf{87.89}	&{--}	&{94.78}	&{89.40}	&{--}	&{99.31} &{54.19}	&--	&{91.74}	&\textbf{23.10}	&--	&61.75	&43.44	&--	&89.31\\
                             \midrule
\multirow{2}{*}{Weakly}                        & \multicolumn{4}{c|}{\multirow{1}{*}{\textbf{Ours ($\lambda=0$)}}} &59.58  &85.74	&90.38	&11.37	&31.94	&44.42	&\textbf{99.66}	&\textbf{100.00}	&\textbf{100.00}	&62.73	&86.53	&90.61	&9.98	&29.67	&41.29	&\textbf{99.99}	&\textbf{100.00}	&\textbf{100.00} \\
                                               & \multicolumn{4}{c|}{\multirow{1}{*}{\textbf{Ours ($\lambda=1$)}}} &{66.07}	&{94.22}	 &97.93	&{16.51}	&49.96	&67.96	&\textbf{99.66}	&\textbf{100.00}	&\textbf100.00	&\textbf{64.74}	&{86.18}	&89.27	&{11.81}	&\textbf{34.77}	&46.83	&\textbf{99.99}	&\textbf{100.00}	&100.00\\

\bottomrule   
\end{tabular}
\label{tab:kitti}
\end{table}

The performance comparison between our method and the state-of-the-art on the KITTI dataset is presented in Tab.~\ref{tab:kitti}. 
It can be seen that among all the comparison algorithms, our method achieves the best cross-area evaluation performance, and the performance discrepancy between same-area and cross-area evaluation of our method is the smallest. 
This is because our method has no information on GT poses, and it is trained to leverage similarity matching for location estimation, preventing itself from overfitting on the GT poses. 
The performance of our method on the same-area evaluation is close to Shi~\etal~\cite{shi2023boosting} and slightly inferior to three recent state-of-the-art methods.
However, in the supplementary material, we demonstrate that the performance of our method on both same-area and cross-area evaluation can be potentially improved when more training image pairs (with pose errors up to tens of meters) are available. 
Our method does not rely on accurate pose labels of training data. 
This leads to significant cost and effort savings by eliminating the need for high-precision pose label acquisition.

The performance comparison between our method and the state-of-the-art on the VIGOR dataset is presented in Tab.~\ref{tab:vigor}. 
The training images in the VIGOR dataset are about $4.5\times$ larger than those in the KITTI dataset, and our method achieves comparable performance with \cite{shi2023boosting} on same-area evaluation. 
Furthermore, similar to the observations on the KITTI dataset, our method's generalization ability from same-area to cross-area is also the best on the VIGOR dataset, with the cross-area evaluation performance surpassing almost all fully supervised methods. Due to the space limit, we provide more discussions in the supplementary material. 

\begin{table}[!t]
    \centering
\setlength{\abovecaptionskip}{0pt}
\setlength{\belowcaptionskip}{0pt}
\setlength{\tabcolsep}{0pt}
\scriptsize 
\caption{\scriptsize Comparison with the state-of-the-art on VIGOR. $*$ indicates full supervision is adopted. 
}
\begin{tabular}{Hc|cc|cc|cc|cc}
\toprule
        \multirow{3}{*}{Supervision} &\multirow{3}{*}{Algorithms} &\multicolumn{4}{c|}{Same-area}  &\multicolumn{4}{c}{\textbf{Cross-area}}  \\ 
        ~ & ~ &\multicolumn{2}{c|}{Aligned-orientation}   &\multicolumn{2}{c|}{Unknown-orientation}   &\multicolumn{2}{c|}{Aligned-orientation}   &\multicolumn{2}{c}{Unknown-orientation}   \\ 
        ~ & ~ & Mean$\downarrow$ & Median$\downarrow $& Mean$\downarrow$ & Median$\downarrow$ & Mean$\downarrow$ & Median$\downarrow$ & Mean$\downarrow$ & Median$\downarrow$ \\ \midrule
        \multirow{4}{*}{Fully} & CVR~\cite{zhu2021vigor}$^*$ & 8.99 & 7.81 & -- & -- & 8.89 & 7.73 & -- & -- \\ 
        ~ & MCC~\cite{xia2022visual}$^*$ & 6.94 & 3.64 & 9.87 & 6.25 & 9.05 & 5.14 & 12.66 & 9.55 \\ 
        ~ & SliceMatch~\cite{lentsch2023slicematch}$^*$  & 5.18 & 2.58 & 8.41 & 5.07 & 5.53 & 2.55 & 8.48 & 5.64 \\ 
        ~ & Shi et al. \cite{shi2023boosting}$^*$ & {4.12} & {1.34} & --& --& 5.16 & \textbf{1.40} & --&-- \\
        ~ & Xia et al. \cite{xia2023convolutional}$^*$ &{3.60} &{1.36} &\textbf{3.74} &\textbf{1.42} & 4.97 &{1.68} & 5.41 & 1.89 \\ 
        ~ & Song et al. \cite{song2024learning}$^*$ &\textbf{3.03} &\textbf{0.97} &{4.97} &{1.90} & 5.01 &{2.42} & 7.67 & 3.67 \\ 
        \midrule
       \multirow{2}{*}{Weakly} &\textbf{Ours} ($\lambda=0$) & 5.22 & 1.97 & 5.33 & 2.09 & 5.37 & 1.93 & 5.37 & 1.93 \\
        ~ &\textbf{Ours} ($\lambda=1$) & {4.19} & {1.68} & 4.18 & 1.66 &{\textbf{4.70}} &{{1.68}} &{\textbf{4.52}} &{\textbf{1.65}} \\ 
   \bottomrule      
\end{tabular}
\label{tab:vigor}
\end{table}



\section{Conclusion}

This paper has introduced the first weakly supervised ground camera pose refinement strategy by ground-to-satellite image registration. 
Given a coarse location and orientation of a ground camera obtained from noisy sensors or visual retrieval techniques, our method is able to refine this pose by ground-to-satellite image registration using a training dataset without accurate pose labels for ground images. 
Key components of our approach include a training scheme for ground-and-satellite rotation alignment using satellite-and-satellite image pairs and a deep metric learning supervision mechanism that trains the network for translation estimation. 
Benefiting from these two innovations, our method, without requiring accurate labels, achieves comparable or superior performance to the recent fully supervised state-of-the-art. 

\section*{Acknowledgements}

This research is funded in part by an ARC Discovery Grant DP220100800.
%
%
\bibliographystyle{splncs04}
\bibliography{egbib}

\begin{thebibliography}{10}
\providecommand{\url}[1]{\texttt{#1}}
\providecommand{\urlprefix}{URL }
\providecommand{\doi}[1]{https://doi.org/#1}

\bibitem{bouazizi2021self}
Bouazizi, A., Wiederer, J., Kressel, U., Belagiannis, V.: Self-supervised 3d
  human pose estimation with multiple-view geometry. In: 2021 16th IEEE
  International Conference on Automatic Face and Gesture Recognition (FG 2021).
  pp.~1--8. IEEE (2021)

\bibitem{Cai_2019_ICCV}
Cai, S., Guo, Y., Khan, S., Hu, J., Wen, G.: Ground-to-aerial image
  geo-localization with a hard exemplar reweighting triplet loss. In: The IEEE
  International Conference on Computer Vision (ICCV) (October 2019)

\bibitem{castaldo2015semantic}
Castaldo, F., Zamir, A., Angst, R., Palmieri, F., Savarese, S.: Semantic
  cross-view matching. In: Proceedings of the IEEE International Conference on
  Computer Vision Workshops. pp. 9--17 (2015)

\bibitem{GeoText1652}
Chu, M., Zheng, Z., Ji, W., Wang, T., Chua, T.S.: Towards natural
  language-guided drones: Geotext-1652 benchmark with spatial relation
  matching. In: Proceedings of the European Conference on Computer Vision
  (ECCV) (2024)

\bibitem{dang2023study}
Dang, T., Kornblith, S., Nguyen, H.T., Chin, P., Khademi, M.: A study on
  self-supervised object detection pretraining. In: Computer Vision--ECCV 2022
  Workshops: Tel Aviv, Israel, October 23--27, 2022, Proceedings, Part IV. pp.
  86--99. Springer (2023)

\bibitem{veronese2015aerial}
De~Paula~Veronese, L., de~Aguiar, E., Nascimento, R.C., Guivant, J.,
  Auat~Cheein, F.A., De~Souza, A.F., Oliveira-Santos, T.: Re-emission and
  satellite aerial maps applied to vehicle localization on urban environments.
  In: 2015 IEEE/RSJ International Conference on Intelligent Robots and Systems
  (IROS). pp. 4285--4290 (2015). \doi{10.1109/IROS.2015.7353984}

\bibitem{fervers2022uncertainty}
Fervers, F., Bullinger, S., Bodensteiner, C., Arens, M., Stiefelhagen, R.:
  Uncertainty-aware vision-based metric cross-view geolocalization. arXiv
  preprint arXiv:2211.12145  (2022)

\bibitem{Geiger2013IJRR}
Geiger, A., Lenz, P., Stiller, C., Urtasun, R.: Vision meets robotics: The
  kitti dataset. International Journal of Robotics Research (IJRR)  (2013)

\bibitem{gharaee2023self}
Gharaee, Z., Lawin, F.J., Forss{\'e}n, P.E.: Self-supervised learning of object
  pose estimation using keypoint prediction. arXiv preprint arXiv:2302.07360
  (2023)

\bibitem{harley2023simple}
Harley, A.W., Fang, Z., Li, J., Ambrus, R., Fragkiadaki, K.: Simple-bev: What
  really matters for multi-sensor bev perception? In: 2023 IEEE International
  Conference on Robotics and Automation (ICRA). pp. 2759--2765. IEEE (2023)

\bibitem{Hu_2018_CVPR}
Hu, S., Feng, M., Nguyen, R.M.H., Hee~Lee, G.: Cvm-net: Cross-view matching
  network for image-based ground-to-aerial geo-localization. In: The IEEE
  Conference on Computer Vision and Pattern Recognition (CVPR) (June 2018)

\bibitem{jaderberg2015spatial}
Jaderberg, M., Simonyan, K., Zisserman, A., et~al.: Spatial transformer
  networks. In: Advances in neural information processing systems. pp.
  2017--2025 (2015)

\bibitem{Lee_2023_CVPR}
Lee, J., Kim, B., Kim, S., Cho, M.: Learning rotation-equivariant features for
  visual correspondence. In: CVPR. pp. 21887--21897 (June 2023)

\bibitem{lentsch2023slicematch}
Lentsch, T., Xia, Z., Caesar, H., Kooij, J.F.: Slicematch: Geometry-guided
  aggregation for cross-view pose estimation. In: Proceedings of the IEEE/CVF
  Conference on Computer Vision and Pattern Recognition. pp. 17225--17234
  (2023)

\bibitem{lin2013cross}
Lin, T.Y., Belongie, S., Hays, J.: Cross-view image geolocalization. In:
  Proceedings of the IEEE Conference on Computer Vision and Pattern
  Recognition. pp. 891--898 (2013)

\bibitem{liu2023self}
Liu, D., Chen, C., Xu, C., Qiu, R.C., Chu, L.: Self-supervised point cloud
  registration with deep versatile descriptors for intelligent driving. IEEE
  Transactions on Intelligent Transportation Systems  (2023)

\bibitem{Liu_2019_CVPR}
Liu, L., Li, H.: Lending orientation to neural networks for cross-view
  geo-localization. In: The IEEE Conference on Computer Vision and Pattern
  Recognition (CVPR) (June 2019)

\bibitem{liu2020high}
Liu, R., Wang, J., Zhang, B.: High definition map for automated driving:
  Overview and analysis. The Journal of Navigation  \textbf{73}(2),  324--341
  (2020)

\bibitem{liu2021swin}
Liu, Z., Lin, Y., Cao, Y., Hu, H., Wei, Y., Zhang, Z., Lin, S., Guo, B.: Swin
  transformer: Hierarchical vision transformer using shifted windows. In:
  Proceedings of the IEEE/CVF International Conference on Computer Vision. pp.
  10012--10022 (2021)

\bibitem{maddern20171}
Maddern, W., Pascoe, G., Linegar, C., Newman, P.: 1 year, 1000 km: The oxford
  robotcar dataset. The International Journal of Robotics Research
  \textbf{36}(1),  3--15 (2017)

\bibitem{mishra2022infra}
Mishra, S., Parchami, A., Corona, E., Chakravarty, P., Vora, A., Parikh, D.,
  Pandey, G.: Localization of a smart infrastructure fisheye camera in a prior
  map for autonomous vehicles. In: 2022 International Conference on Robotics
  and Automation (ICRA). pp. 5998--6004 (2022).
  \doi{10.1109/ICRA46639.2022.9811793}

\bibitem{mousavian2016semantic}
Mousavian, A., Kosecka, J.: Semantic image based geolocation given a map. arXiv
  preprint arXiv:1609.00278  (2016)

\bibitem{pathak2016context}
Pathak, D., Krahenbuhl, P., Donahue, J., Darrell, T., Efros, A.A.: Context
  encoders: Feature learning by inpainting. In: Proceedings of the IEEE
  conference on computer vision and pattern recognition. pp. 2536--2544 (2016)

\bibitem{radford2021learning}
Radford, A., Kim, J.W., Hallacy, C., Ramesh, A., Goh, G., Agarwal, S., Sastry,
  G., Askell, A., Mishkin, P., Clark, J., et~al.: Learning transferable visual
  models from natural language supervision. In: International conference on
  machine learning. pp. 8748--8763. PMLR (2021)

\bibitem{Regmi_2019_ICCV}
Regmi, K., Shah, M.: Bridging the domain gap for ground-to-aerial image
  matching. In: The IEEE International Conference on Computer Vision (ICCV)
  (October 2019)

\bibitem{sarlin2023orienternet}
Sarlin, P.E., DeTone, D., Yang, T.Y., Avetisyan, A., Straub, J., Malisiewicz,
  T., Bul{\`o}, S.R., Newcombe, R., Kontschieder, P., Balntas, V.: Orienternet:
  Visual localization in 2d public maps with neural matching. In: Proceedings
  of the IEEE/CVF Conference on Computer Vision and Pattern Recognition. pp.
  21632--21642 (2023)

\bibitem{shi2022beyond}
Shi, Y., Li, H.: Beyond cross-view image retrieval: Highly accurate vehicle
  localization using satellite image. In: Proceedings of the IEEE/CVF
  Conference on Computer Vision and Pattern Recognition. pp. 17010--17020
  (2022)

\bibitem{shi2019spatial}
Shi, Y., Liu, L., Yu, X., Li, H.: Spatial-aware feature aggregation for image
  based cross-view geo-localization. In: Advances in Neural Information
  Processing Systems. pp. 10090--10100 (2019)

\bibitem{shi2023boosting}
Shi, Y., Wu, F., Perincherry, A., Vora, A., Li, H.: Boosting 3-dof
  ground-to-satellite camera localization accuracy via geometry-guided
  cross-view transformer. arXiv preprint arXiv:2307.08015  (2023)

\bibitem{shi2020looking}
Shi, Y., Yu, X., Campbell, D., Li, H.: Where am {I} looking at? joint location
  and orientation estimation by cross-view matching. In: Proceedings of the
  IEEE/CVF Conference on Computer Vision and Pattern Recognition. pp.
  4064--4072 (2020)

\bibitem{shi2022accurate}
Shi, Y., Yu, X., Liu, L., Campbell, D., Koniusz, P., Li, H.: Accurate 3-dof
  camera geo-localization via ground-to-satellite image matching. IEEE
  Transactions on Pattern Analysis and Machine Intelligence  (2022)

\bibitem{shi2020optimal}
Shi, Y., Yu, X., Liu, L., Zhang, T., Li, H.: Optimal feature transport for
  cross-view image geo-localization. In: AAAI. pp. 11990--11997 (2020)

\bibitem{shi2023cvlnet}
Shi, Y., Yu, X., Wang, S., Li, H.: Cvlnet: Cross-view semantic correspondence
  learning for video-based camera localization. In: Computer Vision--ACCV 2022:
  16th Asian Conference on Computer Vision, Macao, China, December 4--8, 2022,
  Proceedings, Part I. pp. 123--141. Springer (2023)

\bibitem{song2024learning}
Song, Z., Lu, J., Shi, Y., et~al.: Learning dense flow field for
  highly-accurate cross-view camera localization. Advances in Neural
  Information Processing Systems  \textbf{36} (2024)

\bibitem{spurr2021self}
Spurr, A., Dahiya, A., Wang, X., Zhang, X., Hilliges, O.: Self-supervised 3d
  hand pose estimation from monocular rgb via contrastive learning. In:
  Proceedings of the IEEE/CVF International Conference on Computer Vision. pp.
  11230--11239 (2021)

\bibitem{sun2019geocapsnet}
Sun, B., Chen, C., Zhu, Y., Jiang, J.: Geocapsnet: Ground to aerial view image
  geo-localization using capsule network. In: 2019 IEEE International
  Conference on Multimedia and Expo (ICME). pp. 742--747. IEEE (2019)

\bibitem{tang2021self}
Tang, T.Y., De~Martini, D., Wu, S., Newman, P.: Self-supervised learning for
  using overhead imagery as maps in outdoor range sensor localization. The
  International Journal of Robotics Research  \textbf{40}(12-14),  1488--1509
  (2021)

\bibitem{tang2020rsl}
Tang, T.Y., De~Martini, D., Barnes, D., Newman, P.: Rsl-net: Localising in
  satellite images from a radar on the ground. IEEE Robotics and Automation
  Letters  \textbf{5}(2),  1087--1094 (2020)

\bibitem{toker2021coming}
Toker, A., Zhou, Q., Maximov, M., Leal-Taix{\'e}, L.: Coming down to earth:
  Satellite-to-street view synthesis for geo-localization. CVPR  (2021)

\bibitem{vo2016localizing}
Vo, N.N., Hays, J.: Localizing and orienting street views using overhead
  imagery. In: European Conference on Computer Vision. pp. 494--509. Springer
  (2016)

\bibitem{vora2020aerial}
Vora, A., Agarwal, S., Pandey, G., McBride, J.: Aerial imagery based lidar
  localization for autonomous vehicles. arXiv preprint arXiv:2003.11192  (2020)

\bibitem{vyas2022gama}
Vyas, S., Chen, C., Shah, M.: Gama: Cross-view video geo-localization. In:
  Computer Vision--ECCV 2022: 17th European Conference, Tel Aviv, Israel,
  October 23--27, 2022, Proceedings, Part XXXVII. pp. 440--456. Springer (2022)

\bibitem{DeBNet}
Wang, C., Zheng, Z., Ruijie, Q., Yang, Y.: Depth-aware blind image
  decomposition for real-world adverse weather recovery. In: Proceedings of the
  European Conference on Computer Vision (ECCV) (2024)

\bibitem{wang2024fine}
Wang, X., Xu, R., Cui, Z., Wan, Z., Zhang, Y.: Fine-grained cross-view
  geo-localization using a correlation-aware homography estimator. Advances in
  Neural Information Processing Systems  \textbf{36} (2024)

\bibitem{wang2022fully}
Wang, Y., Zhuo, W., Li, Y., Wang, Z., Ju, Q., Zhu, W.: Fully self-supervised
  learning for semantic segmentation. arXiv preprint arXiv:2202.11981  (2022)

\bibitem{workman2015location}
Workman, S., Jacobs, N.: On the location dependence of convolutional neural
  network features. In: Proceedings of the IEEE Conference on Computer Vision
  and Pattern Recognition Workshops. pp. 70--78 (2015)

\bibitem{workman2015wide}
Workman, S., Souvenir, R., Jacobs, N.: Wide-area image geolocalization with
  aerial reference imagery. In: Proceedings of the IEEE International
  Conference on Computer Vision. pp. 3961--3969 (2015)

\bibitem{worrall2017harmonic}
Worrall, D.E., Garbin, S.J., Turmukhambetov, D., Brostow, G.J.: Harmonic
  networks: Deep translation and rotation equivariance. In: Proceedings of the
  IEEE Conference on Computer Vision and Pattern Recognition. pp. 5028--5037
  (2017)

\bibitem{xia2023convolutional}
Xia, Z., Booij, O., Kooij, J.F.: Convolutional cross-view pose estimation.
  arXiv preprint arXiv:2303.05915  (2023)

\bibitem{xia2022visual}
Xia, Z., Booij, O., Manfredi, M., Kooij, J.F.: Visual cross-view metric
  localization with dense uncertainty estimates. In: European Conference on
  Computer Vision. pp. 90--106. Springer (2022)

\bibitem{xia2024adapting}
Xia, Z., Shi, Y., Li, H., Kooij, J.F.: Adapting fine-grained cross-view
  localization to areas without fine ground truth. In: Proceedings of the
  European Conference on Computer Vision (ECCV) (2024)

\bibitem{xiao2020monocular}
Xiao, Z., Yang, D., Wen, T., Jiang, K., Yan, R.: Monocular localization with
  vector hd map (mlvhm): A low-cost method for commercial ivs. Sensors
  \textbf{20}(7), ~1870 (2020)

\bibitem{yang2021cross}
Yang, H., Lu, X., Zhu, Y.: Cross-view geo-localization with layer-to-layer
  transformer. Advances in Neural Information Processing Systems  \textbf{34},
  29009--29020 (2021)

\bibitem{zhai2017predicting}
Zhai, M., Bessinger, Z., Workman, S., Jacobs, N.: Predicting ground-level scene
  layout from aerial imagery. In: IEEE Conference on Computer Vision and
  Pattern Recognition. vol.~3 (2017)

\bibitem{Zhai_2019_ICCV}
Zhai, X., Oliver, A., Kolesnikov, A., Beyer, L.: S4l: Self-supervised
  semi-supervised learning. In: Proceedings of the IEEE/CVF International
  Conference on Computer Vision (ICCV) (October 2019)

\bibitem{zhang2023cross}
Zhang, X., Sultani, W., Wshah, S.: Cross-view image sequence geo-localization.
  In: Proceedings of the IEEE/CVF Winter Conference on Applications of Computer
  Vision. pp. 2914--2923 (2023)

\bibitem{zhang2024increasing}
Zhang, Y., Shi, Y., Wang, S., Vora, A., Perincherry, A., Chen, Y., Li, H.:
  Increasing slam pose accuracy by ground-to-satellite image registration.
  arXiv preprint arXiv:2404.09169  (2024)

\bibitem{Zhu_2022_CVPR}
Zhu, S., Shah, M., Chen, C.: Transgeo: Transformer is all you need for
  cross-view image geo-localization. In: Proceedings of the IEEE/CVF Conference
  on Computer Vision and Pattern Recognition (CVPR). pp. 1162--1171 (June 2022)

\bibitem{zhu2021revisiting}
Zhu, S., Yang, T., Chen, C.: Revisiting street-to-aerial view image
  geo-localization and orientation estimation. In: Proceedings of the IEEE/CVF
  Winter Conference on Applications of Computer Vision. pp. 756--765 (2021)

\bibitem{zhu2021vigor}
Zhu, S., Yang, T., Chen, C.: Vigor: Cross-view image geo-localization beyond
  one-to-one retrieval. CVPR  (2021)

\end{thebibliography}

\newpage
\appendix

\section{Sensitiveness to Coarse Pose Errors}
In this section, we investigate the performance of our method under varying coarse pose errors.

\textbf{Range of location errors.}
Table~\ref{tab:loc} presents the performance comparison between our method and the state-of-the-art [{\color{green}{26, 28}}], across different ranges of location errors: $28 \times 28$ m$^2$ and $56 \times 56$ m$^2$, while maintaining the same orientation ambiguity of $20^\circ$.
The results show that our method achieves consistently the best performance on cross-area evaluation. 


\begin{table}[ht]
\centering
\setlength{\abovecaptionskip}{0pt}
\setlength{\belowcaptionskip}{0pt}
\setlength{\tabcolsep}{5pt}
\scriptsize 
\caption{Performance comparison with different location error ranges on the cross-view KITTI dataset. 
}
\begin{tabular}{c|l|ccH|ccH|ccH}
\toprule
\multirow{3}{*}{\begin{tabular}[c]{@{}c@{}}Location \\ Error ( m$^2$)\end{tabular}} & \multirow{3}{*}{Algorithms} & \multicolumn{9}{c}{Test-1 (Same-area)}                                                                                                                                                                                                                                                                       \\
                                                                                 &                             & \multicolumn{3}{c|}{Lateral}                          & \multicolumn{3}{c|}{Longitudinal}                          & \multicolumn{3}{c}{Azimuth}                                             \\ 
                                                                                &                             & $d=1\uparrow$          & $d=3\uparrow$          & $d=5\uparrow$          & $d=1\uparrow$         & $d=3\uparrow$          & $d=5\uparrow$          & $\theta=1\uparrow$     & $\theta=3\uparrow$     & $\theta=5$         \\ \midrule

\multirow{4}{*}{$28\times28$}                                            & Shi and Li [{\color{green}{26}}]$^*$         & 44.66          & 73.92          & 81.18          & {12.06} & {35.62} & {54.73} & 25.31          & 57.41           & 74.48          \\
                                                                                & Shi \etal [{\color{green}{28}}]$^*$      &  {85.85}   & {98.46}   & {99.55}   & {23.27}     & 46.99     & 58.39    & 98.89      & 99.97      & 100.00       \\ \cmidrule{2-11}
                                                                                & \textbf{Ours ($\lambda=0$)} &61.46 & 87.76 & 92.58 & 13.44 & 38.14 & 54.09 & {99.76} & {100.00 }& {100.00} \\
                                                                                & \textbf{Ours ($\lambda=1$)} & {66.39} & {94.38} & {98.57 }& {18.18 }& {53.59} & {74.53} & {99.76} & {100.00} & {100.00}  \\
                                                                                \midrule
\multirow{4}{*}{$56 \times 56$}                                            & Shi and Li [{\color{green}{26}}]$^*$        & 35.54          & 70.77          & 80.36          & 5.22           & 15.88          & 26.13          & 19.64          & 51.76           & 71.72              \\
                                                                                 & Shi \etal [{\color{green}{28}}]$^*$      & {76.44}	&{96.34}	&{98.89}	&{23.54}	&{50.57}	&{62.18}	&{99.10}	&{100.00}	&{100.00} \\ \cmidrule{2-11}
                                                                                & \textbf{Ours ($\lambda=0$)} &59.58  &85.74	&90.38	&11.37	&31.94	&44.42	&99.66	&100.00	&100.00	 \\
                                                                                & \textbf{Ours ($\lambda=1$)} &66.07	&94.22	 &97.93	&16.51	&49.96	&67.96	&99.66	&100.00	&100.00	\\
                                                                                 \bottomrule
\end{tabular}
\begin{tabular}{c|l|ccH|ccH|ccH}
\toprule
\multirow{3}{*}{\begin{tabular}[c]{@{}c@{}}Location \\ Error ( m$^2$)\end{tabular}} & \multirow{3}{*}{Algorithms}  & \multicolumn{9}{|c}{\textbf{Test-2 (Cross-area)}}                                                                                                                                     \\
                                                                                 &                             & \multicolumn{3}{c|}{Lateral}                          & \multicolumn{3}{c|}{Longitudinal}                          & \multicolumn{3}{c}{Azimuth}                            \\ 
                                                                                &                              & $d=1\uparrow$          & $d=3\uparrow$          & $d=5$          & $d=1\uparrow$         & $d=3\uparrow$          & $d=5$          & $\theta=1\uparrow$     & $\theta=3\uparrow$     & $\theta=5$               \\ \midrule
\multirow{4}{*}{$28\times28$}                                            & Shi and Li [{\color{green}{26}}]$^*$         & 34.17          & 72.30          & 81.15          & 11.56          & {35.08} & {53.77} & 11.40          & 48.18           & 65.80           \\
                                                                                & Shi \etal [{\color{green}{28}}]$^*$          & 60.01   & 87.96   & 92.97   & 14.69     & 35.64     & 48.46    & 99.42      & 100.00     & 100.00  \\ \cmidrule{2-11}
                                                                                & \textbf{Ours ($\lambda=0$)}  & { 65.62} & { 90.32} & { 93.99} & { 13.46} & { 38.53} & { 54.06} & {99.97} & {100.00 }& {100.00} \\
                                                                                & \textbf{Ours ($\lambda=1$)} & {67.90 }& {89.76} & {92.60} & {14.29} & {42.92 }& {58.62} & {99.97} & {100.00} & {100.00} \\
                                                                                \midrule
\multirow{4}{*}{$56 \times 56$}                                            & Shi and Li [{\color{green}{26}}]$^*$       & 27.82          & 59.79          & 72.89          & 5.75           & 16.36          & 26.48          & 18.42          & 49.72           & 71.00           \\
                                                                                 & Shi \etal [{\color{green}{28}}]$^*$      &{57.72}	&{86.77}	&{91.16}	&{14.15}	&{34.59}	&{45.00}	&{98.98}	&{100.00}	&{100.00}\\ \cmidrule{2-11}
                                                                                & \textbf{Ours ($\lambda=0$)} &62.73	&86.53	&90.61	&9.98	&29.67	&41.29	&99.99	&100.00	&100.00 \\
                                                                                & \textbf{Ours ($\lambda=1$)} &64.74	&86.18	&89.27	&11.81	&34.77	&46.83	&99.99	&100.00	&100.00\\
                                                                                 \bottomrule
\end{tabular}

\label{tab:loc}
\end{table}


\begin{table}[t!]
\centering
\setlength{\abovecaptionskip}{0pt}
\setlength{\belowcaptionskip}{0pt}
\setlength{\tabcolsep}{5pt}
\scriptsize 
\caption{Performance comparison with different location error ranges on the cross-view KITTI dataset. 
}
\begin{tabular}{c|c|ccH|ccH|ccH}
\toprule
\multirow{3}{*}{\begin{tabular}[c]{@{}c@{}}Orientation \\ Ambiguity\end{tabular}} & \multirow{3}{*}{Algorithms} & \multicolumn{9}{c|}{Test-1 (Same-area)}     \\
                                                                                 &                             & \multicolumn{3}{c|}{Lateral}                          & \multicolumn{3}{c|}{Longitudinal}                          & \multicolumn{3}{c|}{Azimuth}                              \\ 
                                                                                 &                             & $d=1\uparrow$          & $d=3\uparrow$          & $d=5\uparrow$          & $d=1\uparrow$         & $d=3\uparrow$          & $d=5\uparrow$          & $\theta=1\uparrow$     & $\theta=3\uparrow$     & $\theta=5$      \\ \midrule
\multirow{4}{*}{$20^\circ$}                                                      & Shi and Li [{\color{green}{26}}]$^*$        & 35.54          & 70.77          & 80.36          & 5.22           & 15.88          & 26.13          & 19.64          & 51.76           & 71.72               \\
                                                                                 & Shi \etal [{\color{green}{28}}]$^*$      & {76.44}	&{96.34}	&{98.89}	&{23.54}	&{50.57}	&{62.18}	&{99.10}	&{100.00}	&{100.00} \\ \cmidrule{2-11}
                                                                                & \textbf{Ours ($\lambda=0$)} &59.58  &85.74	&90.38	&11.37	&31.94	&44.42	&99.66	&100.00	&100.00	 \\
                                                                                & \textbf{Ours ($\lambda=1$)} &66.07	&94.22	 &97.93	&16.51	&49.96	&67.96	&99.66	&100.00	&100.00	\\
                                                                                  \midrule
\multirow{4}{*}{$80^\circ$}                                                      & Shi and Li [{\color{green}{26}}]$^*$         & 26.95          & 62.39          & 78.40          & 5.14           & 15.69          & 26.27          & 3.10           & 8.88            & 15.00                  \\
                                                                                 & Shi \etal [{\color{green}{28}}]$^*$      & 70.21   & 95.47   & 98.28   & 22.29     & 48.90     & 59.50    & 53.27      & 93.98      & 98.99     \\ \cmidrule{2-11}
                                                                                & \textbf{Ours ($\lambda=0$)} &53.11 & 86.03 & 90.27 & 12.99 & 32.18 & 42.38 & 57.65 & 96.79 & 99.58 \\
                                                                                & \textbf{Ours ($\lambda=1$)} &57.94 & 91.49 & 96.40 & 17.73 & 47.44 & 62.66 & 57.65 & 96.79 & 99.58 \\ \bottomrule
\end{tabular}
\begin{tabular}{c|c|ccH|ccH|ccH}
\toprule
\multirow{3}{*}{\begin{tabular}[c]{@{}c@{}}Orientation \\ Ambiguity\end{tabular}} & \multirow{3}{*}{Algorithms} & \multicolumn{9}{c}{\textbf{Test-2 (Cross-area)}}                                                                                                                                     \\
                                                                                 &                             & \multicolumn{3}{c|}{Lateral}                          & \multicolumn{3}{c|}{Longitudinal}                          & \multicolumn{3}{c}{Azimuth}                            \\ 
                                                                                 &                             & $d=1\uparrow$          & $d=3\uparrow$          & $d=5$          & $d=1\uparrow$         & $d=3\uparrow$          & $d=5$          & $\theta=1\uparrow$     & $\theta=3\uparrow$     & $\theta=5$               \\ \midrule
\multirow{4}{*}{$20^\circ$}                                                      & Shi and Li [{\color{green}{26}}]$^*$        & 27.82          & 59.79          & 72.89          & 5.75           & 16.36          & 26.48          & 18.42          & 49.72           & 71.00           \\
                                                                                 & Shi \etal [{\color{green}{28}}]$^*$      &{57.72}	&{86.77}	&{91.16}	&{14.15}	&{34.59}	&{45.00}	&{98.98}	&{100.00}	&{100.00}\\ \cmidrule{2-11}
                                                                                & \textbf{Ours ($\lambda=0$)} &62.73	&86.53	&90.61	&9.98	&29.67	&41.29	&99.99	&100.00	&100.00 \\
                                                                                & \textbf{Ours ($\lambda=1$)} &64.74	&86.18	&89.27	&11.81	&34.77	&46.83	&99.99	&100.00	&100.00\\
                                                                                  \midrule
\multirow{4}{*}{$80^\circ$}                                                      & Shi and Li [{\color{green}{26}}]$^*$       & 22.43          & 54.63          & 71.03          & 5.17           & 15.78          & 25.97          & 3.05           & 8.50            & 14.25           \\
                                                                                 & Shi \etal [{\color{green}{28}}]$^*$     & 56.97                     & 87.72                     & 92.35                     & 15.17                     & 35.39                     & 47.02                     & 58.68                          & 95.92                          & 99.15\\ \cmidrule{2-11}
                                                                                & \textbf{Ours ($\lambda=0$)} & 57.68 & 84.92 & 89.59 & 11.64 & 31.52 & 41.16 & 56.79 & 97.76 & 99.71 \\
                                                                                & \textbf{Ours ($\lambda=1$)} & 60.50 & 86.57 & 90.72 & 12.62 & 35.60 & 48.24 & 56.79 & 97.76 & 99.71\\ \bottomrule
\end{tabular}
\label{tab:orien}
\end{table}

\textbf{Variation in orientation ambiguity.}
Subsequently, we augment the orientation ambiguity from $20^\circ$ to $80^\circ$, while maintaining a location error range of $56 \times 56$ m$^2$.
Table~\ref{tab:orien} provides the performance comparison between our method and the two state-of-the-art [{\color{green}{26, 28}}].
Our method achieves the best cross-area evaluation performance on the different orientation ambiguity. 
Furthermore, the results reveal a decline for all methods in the percentage of images for which the estimated orientation is restricted to $1^\circ$ as the orientation ambiguity increases. Nevertheless, our method and Shi \etal [{\color{green}{28}}], which was recently accepted to ICCV2023, consistently maintain the majority of image orientations within a $3^\circ$ margin from their ground truth values. Consequently, the translation estimation performance remains robust.
In contrast, Shi and Li [{\color{green}{26}}] encounter a notable drop in both translation and orientation estimation performance.

\section{Performance with Increasing Amounts of Data as Supervision}

\begin{figure}[t!]
    \centering
    \setlength{\abovecaptionskip}{0pt}
\setlength{\belowcaptionskip}{0pt}
\begin{subfigure}{0.42\linewidth}
    \includegraphics[width=\linewidth]{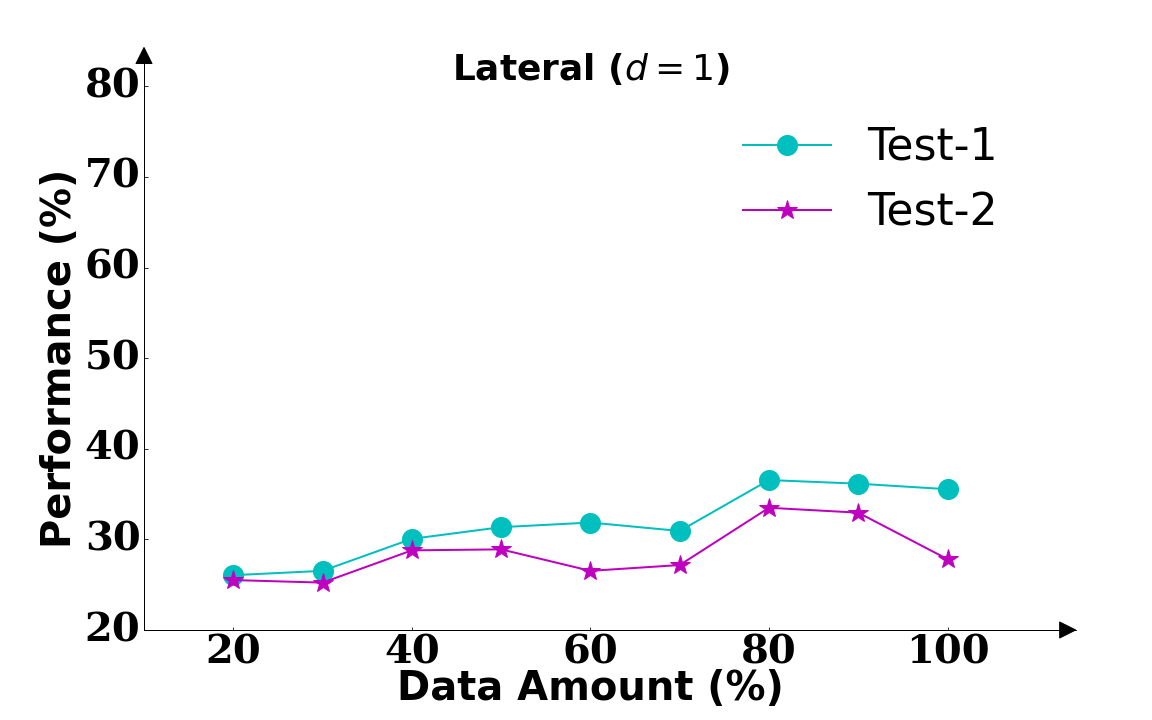}
    \caption{Shi and Li~[{\color{green}{24}}]}
\end{subfigure}
\begin{subfigure}{0.42\linewidth}
    \includegraphics[width=\linewidth]{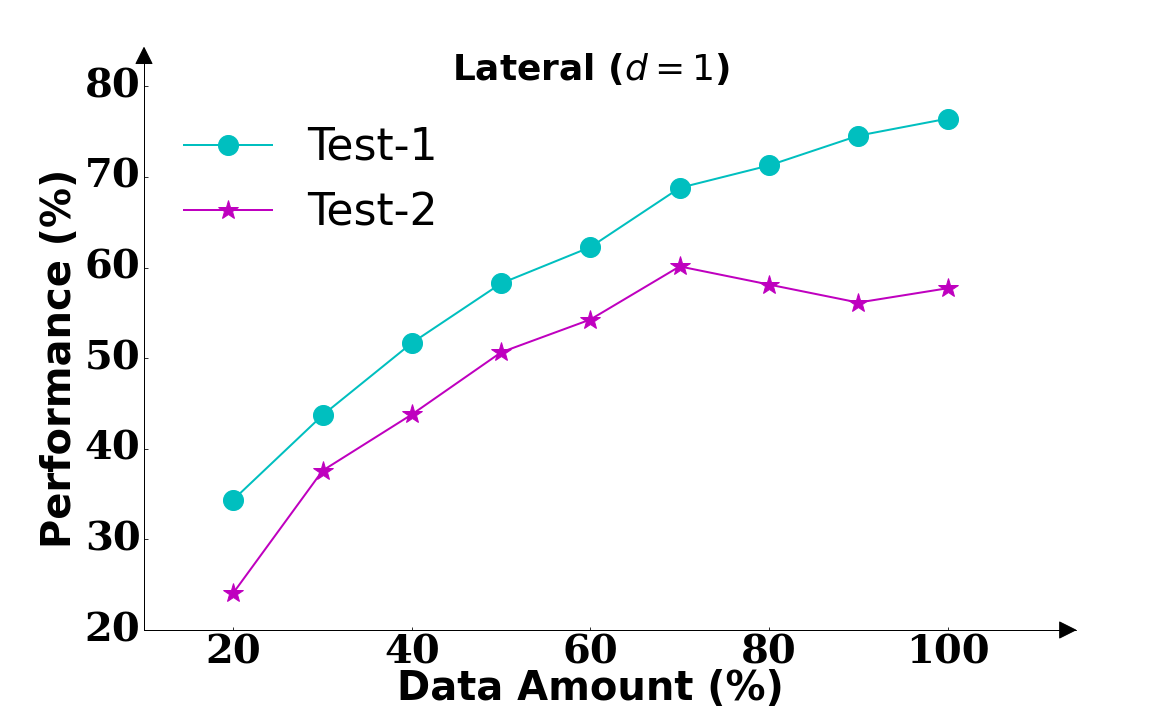}
    \caption{Shi \etal~[{\color{green}{26}}]}
\end{subfigure}
\begin{subfigure}{0.42\linewidth}
    \includegraphics[width=\linewidth]{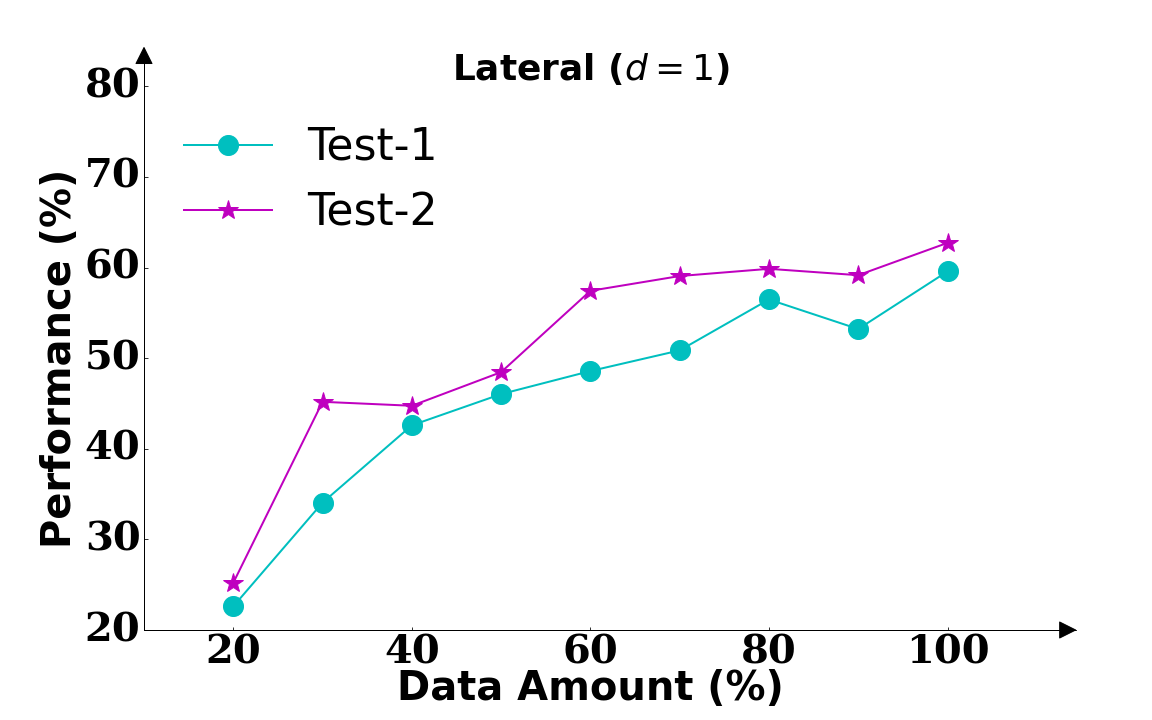}
    \caption{Ours ($\lambda=0$)}
\end{subfigure}
\begin{subfigure}{0.42\linewidth}
    \includegraphics[width=\linewidth]{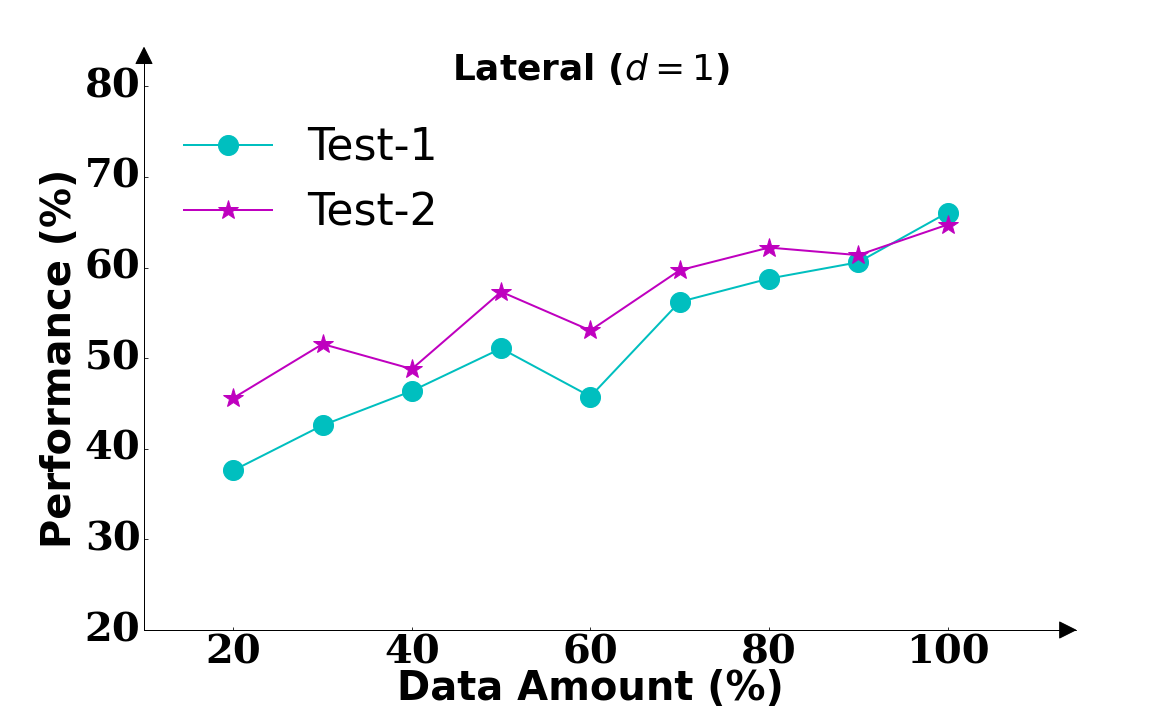}
    \caption{Ours ($\lambda=1$)}
\end{subfigure}
    \caption{Performance comparison between our method and the state-of-the-art on the KITTI dataset with different amounts of training data. }
    \label{fig:amounts}
\end{figure}

Below, we analyze the performance of our method with $\lambda=0, 1$ and the state-of-the-art [{\color{green}{26, 28}}], when different amounts of training data are employed. 
The results are illustrated in Fig.~\ref{fig:amounts}.

For most models, except Shi and Li [{\color{green}{26}}], we observe a consistent increase in performance on the same-area evaluation (Test-1) as the amount of training data increases. However, when it comes to the cross-area evaluation (Test-2), the two state-of-the-art methods, which require ground truth pose for supervision, exhibit a decline in performance when the training data exceeds $80\%$. This phenomenon suggests that our method avoids overfitting and holds the potential for further performance improvements with additional training data. Moreover, it's worth noting that our method doesn't necessitate GT labels for ground images during training, simplifying the process of large-scale data collection and reducing associated costs.

\section{Semi-supervised Setting}
Our method can be easily adapted to address the scenario when a small amount of training data with accurate pose labels is available by adding additional supervision to the network with this amount of data, \eg, using the training objective in Shi \etal~[{\color{green}{28}}].

In Tab.~\ref{tab:semi-supervise}, we show the performance of our method when fine-tuned with different portions of the current training dataset with accurate pose labels. Data amount = 0 indicates the original weakly supervised pipeline where no such data is available.
The results show that fine-tuning leads to better performance when the data amount with accurate pose labels is over 50\%. However, when the data amount is smaller, the fine-tuning leads to inferior performance. We suspect this is because the model overfits the limited training data with accurate labels, causing a loss of general inference ability on other data.

\begin{table*}[t!]
    \centering
    \setlength{\abovecaptionskip}{0pt}
\setlength{\belowcaptionskip}{0pt}
\setlength{\tabcolsep}{3pt}
    \caption{The percentage of images whose lateral translation has been restricted to 1 meter ($d=1$) to its ground truth value under the semi-supervised setting. }
    \begin{tabular}{c|c| c| c| c| c| c| c| c| c| c}
    \toprule
       Data Amount (\%)&0&0.2&0.3&0.4&0.5&0.6&0.7&0.8&0.9 &1 \\ \midrule
Lateral ($d=1$)	&62.73&39.49&44.61&55.31&60.96&65.60&71.48&74.61&78.61&83.86 \\ \bottomrule
    \end{tabular}
    \label{tab:semi-supervise}
\end{table*}

A properly designed training method should address this problem.
However,
apart from this semi-supervised scenario, we believe the original weakly-supervised application addressed in this paper is also of high importance. It avoids additional effort to select the portion of accurate data, as most localization algorithms and noisy GPS sensors themselves do not provide a measure of whether its prediction is accurate or not.
Thus, we leave this semi-supervised setting as a future work.

\section{Sharing Feature Extractors or Not at Different Scenarios}

We empirically found that the feature extractors for satellite and ground images captured by a pin-hole camera are shareable in the localization task. Tab.~\ref{tab:feature_extractors} presents the comparison of our method with shared or non-shared feature descriptors. 
In this comparison, the rotation estimator is kept the same and trained only on satellite images. 
From the results, it can be seen that sharing weights between ground and satellite images achieves better performance. 
A potential explanation might be that both the satellite images and ground images captured by a pin-hole camera map straight lines in the real world to straight lines in images and the viewpoint differences of the two view images are solved by a geometry projection module. This is similar to the task of multi-view stereo and image-based rendering, where the feature extractors for multi-view images are shared, and their differences are handled by Homography/geometry warping. 
Not sharing weights between the two branches increases the learning burden of the network, especially when supervision is not strong, resulting in inferior performance.

While for rotation and translation estimation, we found different feature extractors for different purposes achieve better performance. Tab.~\ref{tab:feature_extractors_RT} illustrates the comparison results. 
This might be because good features for rotation and translation estimation are not identical. When re-using the feature extractors in rotation estimation for translation, we found the network converges slowly, and the performance on both rotation and translation estimation is poor, although better than the original coarse poses that we aim to refine. 

\begin{table}[t!]
\centering
\setlength{\abovecaptionskip}{0pt}
\setlength{\belowcaptionskip}{0pt}
\setlength{\tabcolsep}{0pt}
\scriptsize 
\caption{Sharing feature extractors or not between satellite and ground images captured by a pin-hole camera. 
}
\begin{tabular}{c|c|ccH|ccH|ccH|ccH|ccH|ccH}
\toprule
 \multicolumn{1}{c|}{ \multirow{3}{*}{}}      & \multirow{3}{*}{Share?}                           & \multicolumn{9}{c|}{Test-1 (Same-area)}                                                         & \multicolumn{9}{c}{Test-2 (Cross-area)}                                                         \\
                            \multicolumn{1}{c|}{} &                           & \multicolumn{3}{c|}{Lateral}                          & \multicolumn{3}{c|}{Longitudinal}                         & \multicolumn{3}{c|}{Azimuth}                      & \multicolumn{3}{c|}{Lateral}                          & \multicolumn{3}{c|}{Longitudinal}                         & \multicolumn{3}{c}{Azimuth}                        \\
                             \multicolumn{2}{c|}{}            & $d=1\uparrow$          & $d=3\uparrow$          & $d=5$          & $d=1\uparrow$         & $d=3\uparrow$          & $d=5$          & $\theta=1\uparrow$     & $\theta=3\uparrow$     & $\theta=5$     & $d=1\uparrow$          & $d=3\uparrow$          & $d=5$          & $d=1\uparrow$         & $d=3\uparrow$          & $d=5$          & $\theta=1\uparrow$     & $\theta=3\uparrow$     & $\theta=5$     \\\midrule
 \multirow{2}{*}{\begin{tabular}[c]{@{}c@{}}\textbf{Ours}\\  ($\lambda=0$)\end{tabular}}& No   & 48.64 & 77.37 & 83.12 & 9.97  & 25.63 & 37.13 & 99.66 & 100.00 & 100.00 & 54.49 & 79.75 & 85.79 & 8.96  & 26.54 & 36.75 & 99.99 & 100.00 & 100.00 \\
                                             & \textbf{Yes}  & 59.58 & 85.74 & 90.38 & 11.37 & 31.94 & 44.42 & 99.66 & 100.00 & 100.00 & 62.73 & 86.53 & 90.61 & 9.98  & 29.67 & 41.29 & 99.99 & 100.00 & 100.00 \\ \midrule
\multirow{2}{*}{\begin{tabular}[c]{@{}c@{}}\textbf{Ours}\\  ($\lambda=1$)\end{tabular}} & No   & 62.81 & 93.00 & 98.20 & 19.90 & 55.53 & 72.22 & 99.66 & 100.00 & 100.00 & 62.81 & 84.77 & 87.83 & 13.14 & 36.89 & 48.61 & 99.99 & 100.00 & 100.00 \\
                                             & \textbf{Yes}  & 66.07 & 94.22 & 97.93 & 16.51 & 49.96 & 67.96 & 99.66 & 100.00 & 100.00 & 64.74 & 86.18 & 89.27 & 11.81 & 34.77 & 46.83 & 99.99 & 100.00 & 100.00 \\ 
\bottomrule   
\end{tabular}
\label{tab:feature_extractors}
\end{table}

\begin{table}[h]
\centering
\setlength{\abovecaptionskip}{0pt}
\setlength{\belowcaptionskip}{0pt}
\setlength{\tabcolsep}{0pt}
\scriptsize 
\caption{Sharing feature extractors or not for rotation and translation estimation. 
}
\begin{tabular}{c|c|ccH|ccH|ccH|ccH|ccH|ccH}
\toprule
 \multicolumn{1}{c|}{ \multirow{3}{*}{}}       & \multirow{3}{*}{Share?}                          & \multicolumn{9}{c|}{Test-1 (Same-area)}                                                         & \multicolumn{9}{c}{Test-2 (Cross-area)}                                                         \\
                            \multicolumn{1}{c|}{} &                           & \multicolumn{3}{c|}{Lateral}                          & \multicolumn{3}{c|}{Longitudinal}                         & \multicolumn{3}{c|}{Azimuth}                      & \multicolumn{3}{c|}{Lateral}                          & \multicolumn{3}{c|}{Longitudinal}                         & \multicolumn{3}{c}{Azimuth}                        \\
                             \multicolumn{2}{c|}{}            & $d=1\uparrow$          & $d=3\uparrow$          & $d=5$          & $d=1\uparrow$         & $d=3\uparrow$          & $d=5$          & $\theta=1\uparrow$     & $\theta=3\uparrow$     & $\theta=5$     & $d=1\uparrow$          & $d=3\uparrow$          & $d=5$          & $d=1\uparrow$         & $d=3\uparrow$          & $d=5$          & $\theta=1\uparrow$     & $\theta=3\uparrow$     & $\theta=5$     \\\midrule
\multirow{2}{*}{\begin{tabular}[c]{@{}c@{}}\textbf{Ours}\\  ($\lambda=0$)\end{tabular}}& Yes    & 33.77 & 74.66 & 85.63 & 9.17  & 26.13 & 37.64 & 11.69 & 34.64  & 54.89  & 32.18 & 71.92 & 82.54 & 7.58  & 24.00 & 34.88 & 12.64 & 37.24  & 59.08  \\
                                            & \textbf{No} & 59.58 & 85.74 & 90.38 & 11.37 & 31.94 & 44.42 & 99.66 & 100.00 & 100.00 & 62.73 & 86.53 & 90.61 & 9.98  & 29.67 & 41.29 & 99.99 & 100.00 & 100.00 \\ \midrule
\multirow{2}{*}{\begin{tabular}[c]{@{}c@{}}\textbf{Ours}\\  ($\lambda=1$)\end{tabular}} & Yes & 35.62 & 81.69 & 92.71 & 10.68 & 31.78 & 46.46 & 10.20 & 31.35  & 52.74  & 32.62 & 73.72 & 84.51 & 8.87  & 26.07 & 37.91 & 10.21 & 31.62  & 50.91  \\
                                             & \textbf{No}  & 66.07 & 94.22 & 97.93 & 16.51 & 49.96 & 67.96 & 99.66 & 100.00 & 100.00 & 64.74 & 86.18 & 89.27 & 11.81 & 34.77 & 46.83 & 99.99 & 100.00 & 100.00 \\
\bottomrule   
\end{tabular}
\label{tab:feature_extractors_RT}
\end{table}

\begin{figure}[t]]
\centering
  \centering
  \includegraphics[width=\linewidth]{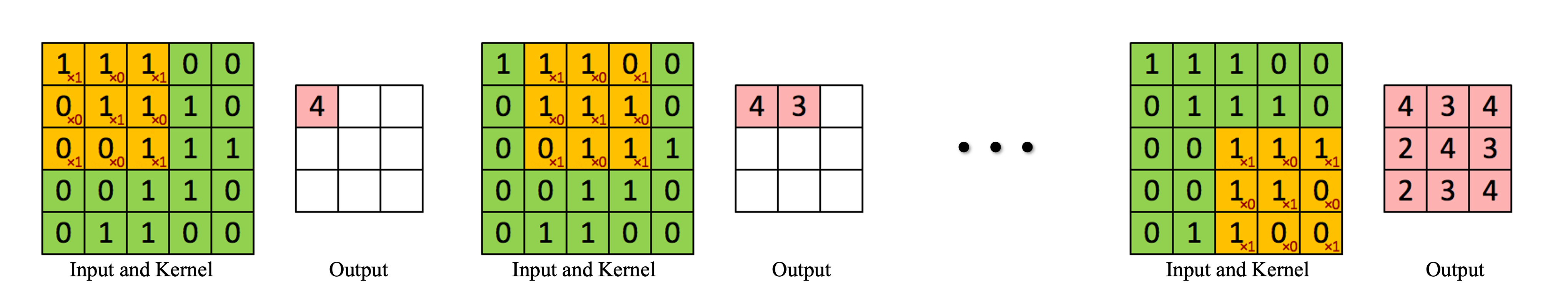} 
  \caption{The spatial correlation process. We compute the inner product between the reference satellite feature map (input) and synthesized overhead view feature map (kernel) from the ground image across all possible locations. This figure is from \url{https://giphy.com/gifs/blog-daniel-keypoints-i4NjAwytgIRDW}}
    \label{fig:convolution}
\end{figure}

\section{Performance on the Ford Dataset}
We present the performance of our method on the Ford dataset in Tab.~\ref{tab:ford}. 
The results show our method achieves competitive performance with fully supervised SOTA, and sharing feature extractors between the ground and satellite branches performs better than not sharing, which is consistent with our observations on KITTI. 

\begin{table}[h!]
\centering
\setlength{\abovecaptionskip}{0pt}
\setlength{\belowcaptionskip}{0pt}
\setlength{\tabcolsep}{6pt}
\scriptsize 
\caption{Average results comparison on Log1 and Log2 of the Ford Multi-AV dataset.}
\begin{tabular}{c|c|cc|cc|cc}
\hline
\multicolumn{2}{c|}{\multirow{2}{*}{Algorithms}}  & \multicolumn{2}{c|}{Lateral}                      & \multicolumn{2}{c|}{Longitudinal}                 & \multicolumn{2}{c}{Azimuth}                      \\
\multicolumn{2}{c|}{}       & $d=1\uparrow$          & $d=3\uparrow$                & $d=1\uparrow$          & $d=3\uparrow$              & $\theta=1\uparrow$     & $\theta=3\uparrow$        \\ \hline
\multicolumn{2}{c|}{Shi and Li~[{\color{green}{26}}]*    }                      &  37.29          & 69.33          & 5.22           & 15.94          & 16.87          & 44.14                 \\
\multicolumn{2}{c|}{Shi \etal [{\color{green}{28}}]*}  & 63.77          & 82.53          & \textbf{19.45} & 34.17          & \textbf{61.49} & 92.62   \\ 
\multicolumn{2}{c|}{Song \etal [{\color{green}{33}}]*}  & 46.13          & -              & 11.97          & -              & 58.64          & -        \\
\hline
\multirow{2}{*}{ \textbf{Ours} ($\lambda=0$)}     & share                    & 65.47          & 93.50          & 8.00           & 20.11          & 46.63          & \textbf{96.86}                  \\
                                                  & not share                    &   14.21          & 69.62          & 5.13           & 13.40          & 46.63          & \textbf{96.86} \\ \hline
\multirow{2}{*}{ \textbf{Ours} ($\lambda=1$) }   & share                       & \textbf{78.87} & \textbf{97.96} & 16.42          & \textbf{37.26} & 46.63          & \textbf{96.86}                \\
                                                 & not share                       & 58.19          & 96.83          & 10.14          & 25.33          & 46.63          & \textbf{96.86}             \\
\hline
\end{tabular}
\label{tab:ford}
\vspace{-1.5em}
\end{table}

\section{Visual Explanation of the Spatial Correlation Process}
The spatial correlation process is 
illiustrated in Fig.~\ref{fig:convolution}. 
We first center crop the synthesized overhead view feature map depicted in Fig.~{\color{red}{5}} (b) of the main paper and make its coverage around $40 \times 40$ m$^2$, as we consider scene contents within $20$m to the camera location is the most important for the localization purpose. 
Then, we adopt the cropped overhead view feature map as the correlation kernel, similar to the yellow kernel in Fig.~\ref{fig:convolution}, and the reference satellite image as the correlation input, indicated by the green grid in Fig.~\ref{fig:convolution}, and apply inner product between the input and the kernal. 
The output, indicated by the pink grid in Fig.~\ref{fig:convolution}, is the location probability map of the ground camera with respect to the satellite image. 

In practice, the coverage of the reference satellite map and the kernel is engineered to make the coverage of the convolution output slightly larger than the location search space of the ground camera. 
In this example, the coverage of the satellite map is around $100 \times 100$ m$^2$, and that of the convolution output (location probability map) is about $60 \times 60$ m$^2$.

\begin{table}[t]
\centering
\setlength{\abovecaptionskip}{0pt}
\setlength{\belowcaptionskip}{0pt}
\setlength{\tabcolsep}{0pt}
\scriptsize 
\caption{Comparison between projecting features (Feat.) and images (Imgs). 
}
\begin{tabular}{c|c|ccH|ccH|ccH|ccH|ccH|ccH}
\toprule
 \multicolumn{2}{c|}{ \multirow{3}{*}{}}                                 & \multicolumn{9}{c|}{Test-1 (Same-area)}                                                         & \multicolumn{9}{c}{Test-2 (Cross-area)}                                                         \\
                            \multicolumn{2}{c|}{}                           & \multicolumn{3}{c|}{Lateral}                          & \multicolumn{3}{c|}{Longitudinal}                         & \multicolumn{3}{c|}{Azimuth}                      & \multicolumn{3}{c|}{Lateral}                          & \multicolumn{3}{c|}{Longitudinal}                         & \multicolumn{3}{c}{Azimuth}                        \\
                             \multicolumn{2}{c|}{}            & $d=1\uparrow$          & $d=3\uparrow$          & $d=5$          & $d=1\uparrow$         & $d=3\uparrow$          & $d=5$          & $\theta=1\uparrow$     & $\theta=3\uparrow$     & $\theta=5$     & $d=1\uparrow$          & $d=3\uparrow$          & $d=5$          & $d=1\uparrow$         & $d=3\uparrow$          & $d=5$          & $\theta=1\uparrow$     & $\theta=3\uparrow$     & $\theta=5$     \\\midrule
  \multirow{2}{*}{\begin{tabular}[c]{@{}c@{}}\textbf{Ours}\\  ($\lambda=0$)\end{tabular}} & Imgs   & 45.40 & 80.57 & 86.75 & 6.97  & 21.39 & 33.98 & 99.92 & 100.00 & 100.00 & 48.24 & 79.16 & 85.44 & 6.95  & 20.60 & 32.91 & 100.00 & 100.00 & 100.00 \\
                                              & \textbf{Feat} & 59.58 & 85.74 & 90.38 & 11.37 & 31.94 & 44.42 & 99.66 & 100.00 & 100.00 & 62.73 & 86.53 & 90.61 & 9.98  & 29.67 & 41.29 & 99.99  & 100.00 & 100.00 \\ \midrule
 \multirow{2}{*}{\begin{tabular}[c]{@{}c@{}}\textbf{Ours}\\  ($\lambda=1$)\end{tabular}} & Imgs   & 54.31 & 90.59 & 96.79 & 13.36 & 38.32 & 56.75 & 99.92 & 100.00 & 100.00 & 58.53 & 87.34 & 91.85 & 11.24 & 33.23 & 47.14 & 100.00 & 100.00 & 100.00 \\
                                              & \textbf{Feat.} & 66.07 & 94.22 & 97.93 & 16.51 & 49.96 & 67.96 & 99.66 & 100.00 & 100.00 & 64.74 & 86.18 & 89.27 & 11.81 & 34.77 & 46.83 & 99.99  & 100.00 & 100.00 \\
\bottomrule   
\end{tabular}
\label{tab:featVSimg}
\end{table}

\section{Stage-wise Training VS. End-to-end Training}

We present the comparison between end-to-end and stage-wise training in Tab.~\ref{tab:e2e}. We observe that, at the beginning of end-to-end training, the rotation estimation performance for the query image is poor (near random), which negatively impacts the translation estimation performance. 
At the same time, allowing the signal of translation estimation loss to propagate back to the rotation estimator negatively affects its performance. 
Furthermore, end-to-end training requires large GPU memory and, thus, a smaller batch size, which is detrimental to metric learning performance. 

\begin{table}[h]
\setlength{\abovecaptionskip}{0pt}
\setlength{\belowcaptionskip}{0pt}
\setlength{\tabcolsep}{8pt}
\centering
\scriptsize
\caption{Comparison with end-to-end training on KITTI dataset (Ours with $\lambda=0$)}. 
\begin{tabular}{c|ccc|ccc}
\hline
\multirow{3}{*}{\begin{tabular}[c]{@{}c@{}}Training \\ Approach\end{tabular}} & \multicolumn{3}{c|}{Test-1 (Same-area)}                                                                                                                   & \multicolumn{3}{c}{Test-2 (Cross-area)}                                                                                                                  \\
                                                                              & \begin{tabular}[c]{@{}c@{}}Lateral \\ $d=1$\end{tabular} & \begin{tabular}[c]{@{}c@{}}Longitudinal\\ $d=1$\end{tabular} & \begin{tabular}[c]{@{}c@{}}Azimuth\\ $\theta=1^\circ$\end{tabular} & \begin{tabular}[c]{@{}c@{}}Lateral \\ $d=1$\end{tabular} & \begin{tabular}[c]{@{}c@{}}Longitudinal\\ $d=1$\end{tabular} & \begin{tabular}[c]{@{}c@{}}Azimuth\\ $\theta=1^\circ$\end{tabular} \\ \hline

End-to-end                                                                    & 32.83                                           &  8.22                                          &  10.15                                             &  32.53                                           &  7.76                                         & 10.10         \\ 
Stage-wise                                                                    & \textbf{59.58 }                                           & \textbf{11.37 }                                          & \textbf{99.66}                                               & \textbf{62.73}                                            & \textbf{9.98 }                                           & \textbf{99.99}                                               \\ \hline                                 
\end{tabular}
\label{tab:e2e}
\vspace{-1em}
\end{table}

\section{Angle ambiguity at $0^\circ \&$  $360^\circ$, why not rotation matrix?}
Since we have prior knowledge of the coarse orientation, the angle ambiguity can be restricted to be smaller than $360^\circ$. Thus, the angle ambiguity at $0^\circ$ and $360^\circ$ can be avoided. 
A rotation matrix over-parameterizes the 1-DoF rotation angle in the cross-view image matching, thus leading to inferior performance, as shown in the 1st row of Tab.~\ref{tab:rot_abla}. 
We further make the pose regressor output sine and cosine of the angle (2-DoF output). The results in the 2nd row of Tab.~\ref{tab:rot_abla} are promising but still inferior to our original parameterization (last row).

\begin{table}[h]
\setlength{\abovecaptionskip}{0pt}
\setlength{\belowcaptionskip}{0pt}
\setlength{\tabcolsep}{15pt}
\centering
\scriptsize
\caption{Additional ablation study on rotation estimation on the KITTI dataset.}
\begin{tabular}{c|c|cc|cc}
\hline
                       \multicolumn{2}{c|}{\multirow{2}{*}{Rotation Parameterization}}                                 & \multicolumn{2}{c}{Test-1 (Same-area)}          & \multicolumn{2}{c}{Test-2 (Cross-area)}          \\
                       \multicolumn{2}{c|}{}                               & $\theta=1^\circ$ & $\theta=3^\circ$ & $\theta=1^\circ$ & $\theta=3^\circ$ \\\hline
\multicolumn{2}{c|}{Rotation Matrix}                    & 5.05             & 15.44            & 5.07             & 15.51          \\
\multicolumn{2}{c|}{Sin(angle) \& Cos(angle)}                    & 90.01             & \textbf{100.00}            & 86.52             & \textbf{100.00}          \\ \hline  
\multicolumn{2}{c|}{Angle}  & \textbf{99.66 }           & \textbf{100.00}              & \textbf{99.99}            & \textbf{100.00}              \\
                       \hline
\end{tabular}
\label{tab:rot_abla}
\end{table}

\section{Comparison Between Projecting Features and Images}
In this paper, we follow the general practice of projecting features instead of images [{\color{green}{32}}]. This is because when projecting ground images to an overhead view by assuming ground plane Homography, the pixels for scene objects above the ground plane are incorrectly projected to the overhead view and thus suffer distortion. In this way, the scene information of these objects will be lost in the projected image, resulting in inferior localization performance. 

In contrast, features have a larger field of view of the original image and encode higher-level semantic information about the scene. For example, the building roots also have a semantic meaning of ``building''. It can be mapped to the building roof in the overhead view, which shares the same semantic information as the building root. Thus, projecting features instead of the original images can tolerate the errors in the overhead-view projection by the ground plane Homography to some extent. 
We illustrate the experimental comparison between projecting features and images in Tab.~\ref{tab:featVSimg}. Not surprisingly, projecting features achieves better performance.

\begin{table}[!t]
    \centering
    \setlength{\abovecaptionskip}{0pt}
\setlength{\belowcaptionskip}{0pt}
\setlength{\tabcolsep}{8pt}
\scriptsize 
\caption{\scriptsize Model size and evaluation speed comparison on the KITTI dataset. 
}
    \begin{tabular}{c|c|c|c|c|c}
    \toprule
        \multicolumn{3}{c|}{Model Size} &\multicolumn{3}{c}{Evaluation Speed} \\ \midrule
        Shi and Li [{\color{green}{26}}] & Shi \etal [{\color{green}{28}}] & \textbf{Ours}  & Shi and Li [{\color{green}{26}}] & Shi at al. [{\color{green}{28}}] & \textbf{Ours}  \\ \midrule
         20.2 M& 29.1 M& 20.6 M & 500 ms & 200 ms& 47 ms \\
         \bottomrule
    \end{tabular}
    \label{tab:complexity}
\end{table}

\section{Model Size and Evaluation Speed Comparison}

We present the model size and evaluation speed comparison with two recent state-of-the-art, whose models and evaluation scripts have been released, in Tab.~\ref{tab:complexity}. All of them are evaluated on an RTX 3090 GPU. It can be seen that our method achieves the fastest evaluation speed with a relatively small model size.

\section{Limitations}

Although our self-supervised learning approach has achieved promising results, it has a few limitations. 

\textbf{(i)} First, as explained previously, our self-supervised training strategy for rotation estimation is only suitable for ground images captured by a {pin-hole} camera. 
Due to the significant domain differences between panoramic and satellite images, it cannot be applied to estimating a {spherical} camera's orientation. 

\textbf{(ii)} Second, our deep metric learning supervision strategy computes the spatial correlation between each query image and several satellite images. To save GPU memory and enable a reasonable batch size for metric learning, we use the feature level of a quarter of the original image size for the translation estimation. 
This actually sacrifices localization accuracy to some extent. 

\textbf{(iii)} For the same reason, we cannot adopt a more sophisticated overhead-view feature synthesis method because it will consume significant memory, thus sacrificing batch size, and the weak supervision limits the learning ability of a powerful overhead-view synthesis module with complex designs. 

\begin{figure}[h]
\vspace{-1em}
\setlength{\abovecaptionskip}{0pt}
\setlength{\belowcaptionskip}{0pt}
    \centering
        \includegraphics[width=0.33\linewidth, height=0.12\linewidth]{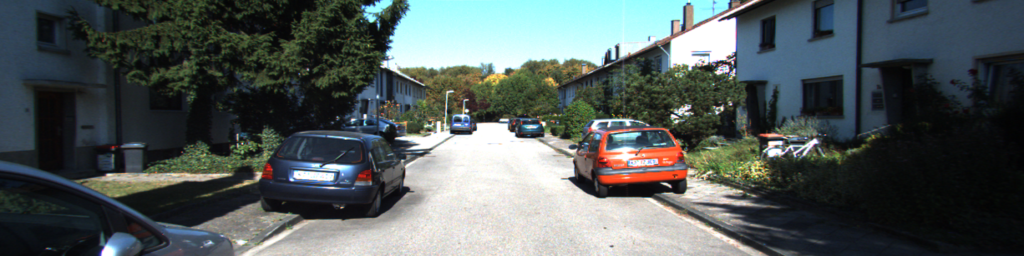}
        \includegraphics[width=0.14\linewidth]{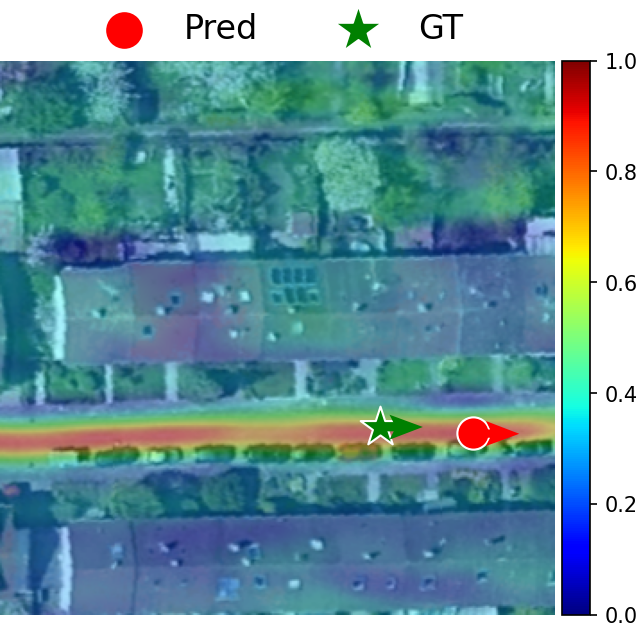}
        \includegraphics[width=0.33\linewidth, height=0.12\linewidth]{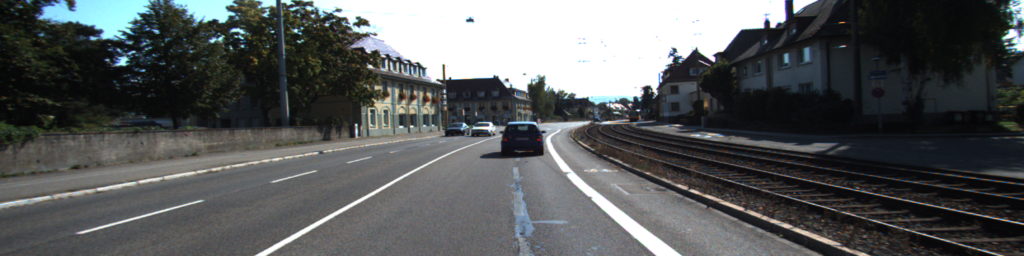}
        \includegraphics[width=0.14\linewidth]{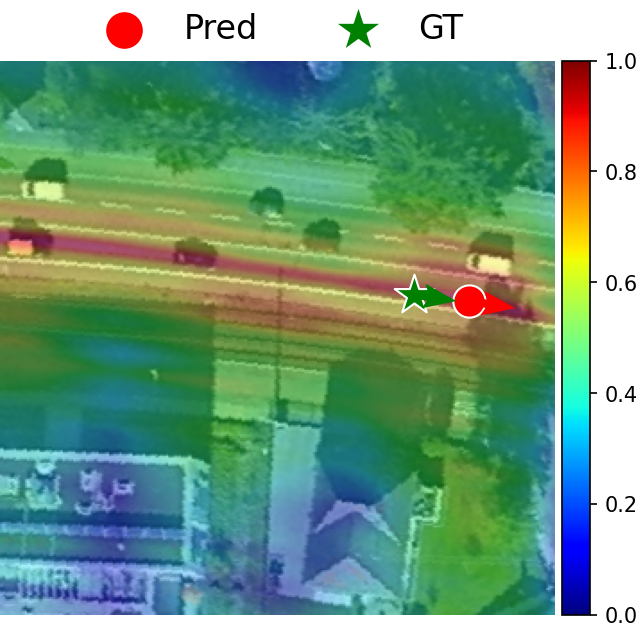}
    \caption{Ambiguity along the longitudinal (driving direction) pose estimation. }
    \label{fig:enter-label}
\end{figure}

\textbf{(vi)} Finally, similar to all the ground-to-satellite localization networks where a single camera is used for query, our method suffers poor localization performance along the longitudinal direction, as shown in Fig.~\ref{fig:enter-label}. 
This can potentially be addressed using a video or multi-camera setup for the query.

We leave these unsolved problems as our future work and encourage the community to pay attention to them.

\section{Potential Negative Impact}

This paper introduces a novel approach for self-supervised ground-to-satellite image registration. The objective of this approach is to enhance the accuracy of coarse camera pose estimates, such as those obtained from noisy GPS sensors, through ground-to-satellite image matching.
However, it also raises concerns about privacy, particularly regarding the potential for individuals or sensitive locations to be identified and tracked without their consent. Unauthorized access to satellite imagery could enable surveillance activities or intrusions into personal privacy, raising ethical and legal implications.



We emphasize that the proposed method should be utilized in a manner that aligns with legal and ethical considerations. Careful implementation and adherence to privacy regulations and policies are crucial to ensure the ethical and responsible use of this approach.

\end{document}